\documentclass[letterpaper, 10 pt, conference]{ieeeconf}  

\usepackage[utf8]{inputenc}
\usepackage[usenames, dvipsnames]{color}
\usepackage{graphicx}
\usepackage{enumerate}
\usepackage{multirow}
\usepackage{graphics}
\usepackage{svg}
\usepackage{graphicx}

\usepackage{amsmath}
\usepackage{nth}
\usepackage{makecell}
\usepackage{colortbl}

\usepackage{caption}
\usepackage{tabularx}

\usepackage{tikz}
\usetikzlibrary{shapes,arrows}
\usepackage{subcaption}
\usepackage{amsmath,amsthm,amssymb,mathrsfs} 
\usepackage[final]{pdfpages}
\usepackage[]{placeins,flafter}
\usepackage[none]{hyphenat} 
\usepackage{xcolor}
\usepackage{adjustbox}
\usepackage{balance}
\usepackage{comment}
\usepackage{nicefrac}
\usepackage{url}
\usepackage{hyperref}
\usepackage{float}
\usepackage{multicol}

\IEEEoverridecommandlockouts                              

\title{\LARGE \bf
Jacobian Computation for Cumulative B-Splines on SE(3) and Application to Continuous-Time Object Tracking
}

\author{Javier Tirado and Javier Civera
\thanks{Javier Tirado and Javier Civera are with I3A, Universidad de Zaragoza, Spain
        {\tt\small jcivera@unizar.es, jtiradogarin@gmail.com }}%
}

\usepackage{lipsum}
\usepackage{empheq}
\usepackage{wrapfig}

\newcommand{\tikzcircle}[2][red,fill=red]{\tikz[baseline=-0.5ex]\draw[#1,radius=#2] (0,0) circle ;}%
\definecolor{blizzardblue}{rgb}{0.67, 0.9, 0.93}

\newcommand{\red}[1]{#1} 
\newcommand{\redd}[1]{#1} 

\newcommand{\Rsix}{\mathbb{R}^6}

\newcommand{\bR}{\mathbf{R}}
\newcommand{\bT}{\mathbf{T}}
\newcommand{\bt}{\mathbf{t}}
\newcommand{\bp}{\mathbf{p}}
\newcommand{\bI}{\mathbf{I}}
\newcommand{\bx}{\mathbf{x}}

\newcommand{\be}{\mathbf{e}}
\newcommand{\bzero}{\mathbf{0}}

\newcommand{\bg}{\mathbf{g}}
\newcommand{\bH}{\mathbf{H}}
\newcommand{\bJ}{\mathbf{J}}
\newcommand{\bSigma}{\boldsymbol{\Sigma}}
\newcommand{\bA}{\mathbf{A}}
\newcommand{\ba}{\mathbf{a}}
\newcommand{\bP}{\mathbf{P}}
\newcommand{\bN}{\mathbf{N}}
\newcommand{\bC}{\mathbf{C}}
\newcommand{\bG}{\mathbf{G}}

\newcommand{\bomega}{\boldsymbol{\omega}}
\newcommand{\bomegahat}{\boldsymbol{\omega}^{\wedge}}

\newcommand{\bv}{\mathbf{v}}

\newcommand{\bOmega}{\boldsymbol{\Omega}}
\newcommand{\bxi}{\boldsymbol{\xi}}
\newcommand{\pt}{\partial}
\newcommand{\proj}{\text{proj}}
\newcommand{\vect}{\text{vec}}

\newcommand{\btau}{\boldsymbol{\tau}}
\newcommand{\btauhat}{\boldsymbol{\tau}^{\wedge}}

\newcommand{\Exp}{\text{Exp}}
\newcommand{\Log}{\text{Log}}

\newcommand{\bJl}{\mathbf{J}_l}

\begin{document}

\maketitle
\thispagestyle{empty}
\pagestyle{empty}

\begin{abstract}

In this paper we propose a method that estimates the $SE(3)$ continuous trajectories (orientation and translation) of the dynamic rigid objects present in a scene, from multiple RGB-D views. Specifically, we fit the object trajectories to cumulative B-Splines curves, which allow us to interpolate, at any intermediate time stamp, not only their poses but also their linear and angular velocities and accelerations. Additionally, we derive in this work the analytical $SE(3)$ Jacobians needed by the optimization, being applicable to any other approach that uses this type of curves. To the best of our knowledge this is the first work that proposes 6-DoF continuous-time object tracking, which we endorse with significant computational cost reduction thanks to our analytical derivations. We evaluate our proposal in synthetic data and in a public benchmark, showing competitive results in localization and significant improvements in velocity estimation in comparison to discrete-time approaches.
\end{abstract}


\section{INTRODUCTION}

\begin{figure}[t]
    \centering
    \begin{subfigure}[t]{0.49\linewidth}
      \centering
        \includegraphics[width=1.0\textwidth]{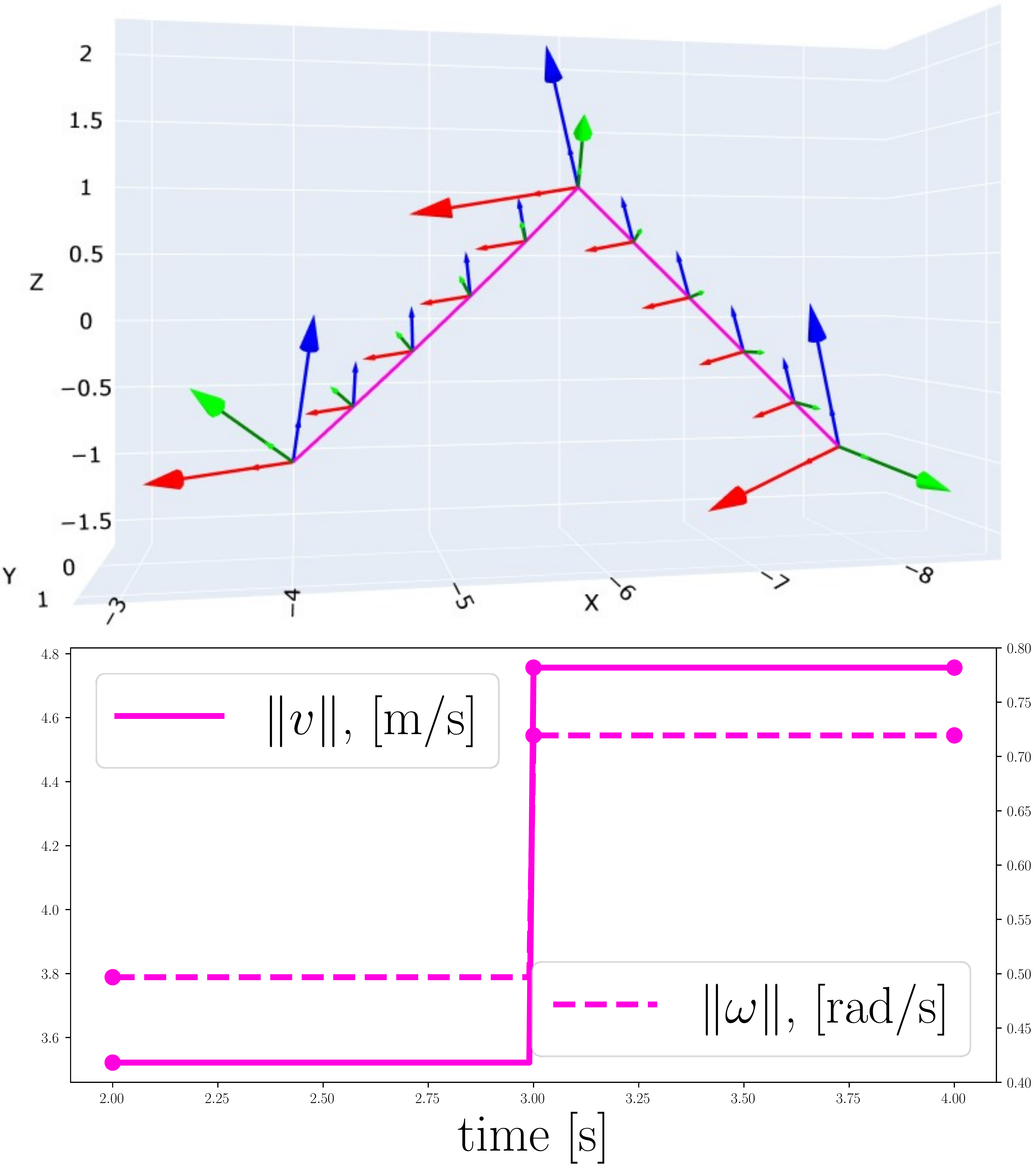}
      \caption{Discrete-time}
    \end{subfigure}
    \hfill
    \begin{subfigure}[t]{0.49\linewidth}
      \centering
      \includegraphics[width=1.0\textwidth]{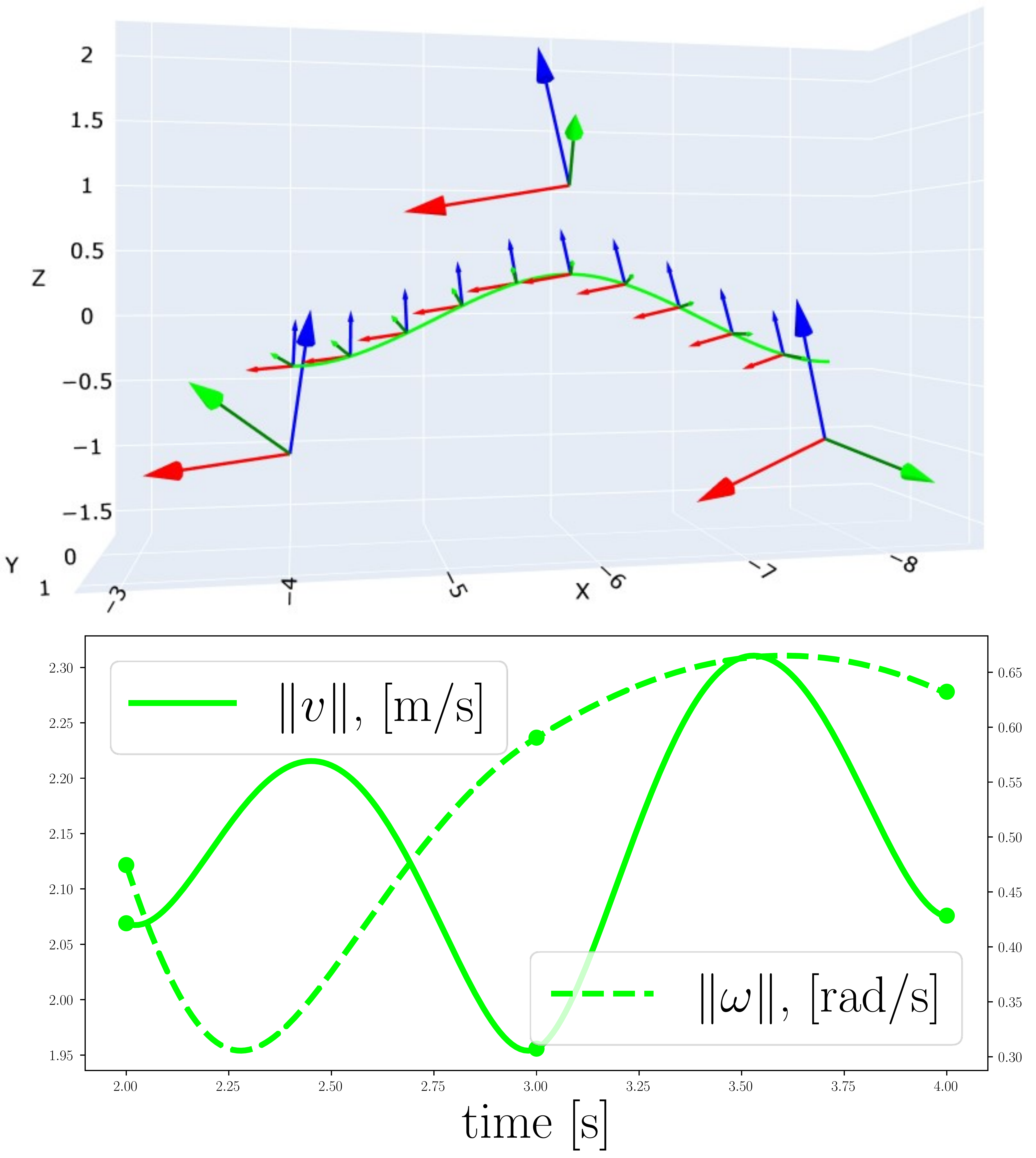}
      \caption{Continuous-time}
    \end{subfigure}
    \caption{Example of linear and angular speed profiles and interpolated poses for discrete-time (a) and continuous-time (b) formulations. In discrete time, the common assumption of constant velocity, leads to discontinuities in time domain. As can be seen, this does not happen with the continuous-time formulation. We also ensure time continuity in acceleration.}
    \label{fig:fig1}
    \vspace{-0.5cm}
\end{figure}

Understanding the dynamic behavior of the moving elements present in a scene can become crucial in several robotic applications. Specifically, in SLAM \cite{cadena2016past} and SfM \cite{schonberger2016structure}, it is well known that models unaware of the non-rigidity of the scene lead to poor performances regarding robustness and accuracy in localization and map reconstruction tasks \cite{saputra2018visual}. Dynamic scenes are indeed acknowledged as a research challenge and several works have addressed such problem by developing systems that first detect, and subsequently remove from the map, the dynamic regions of an image \cite{bescos2018dynaslam, kim2016effective, ballester2021dot}.

On the other hand, the challenge of estimating the unconstrained motions of dynamic objects (instead of removing them) has recently gained attention. Acknowledging the kinematics of moving objects is expected to benefit the reasoning of the agents over the scene, increasing their capabilities for decision making. Moreover, this information can be used to enhance AR/VR experiences or serve as an alternative to expensive motion-capture set-ups. In this paper we focus on rigid objects with a free 6-degrees-of-freedom (6-DoF) motion.

With these goals, we find in the literature several approaches that aim to estimate the position and orientation of the objects ($SE(3)$ poses) \cite{judd2018multimotion, zhang2020vdo, huang2020clustervo}. However, relevant kinematic magnitudes such as the linear and angular velocities are not considered in their models. Thereby the estimates do not need to explicitly follow physically feasible motions. More recent works \cite{bescos2021dynaslam, judd2021multimotion} assume that kinematics between consecutive images are similar. Imposing this constraint, they are able to not only estimate the $SE(3)$ pose but also velocities \cite{bescos2021dynaslam} and accelerations \cite{judd2021multimotion}.

All these works operate on discrete time, returning estimations at fixed time stamps. Thereby the estimated kinematics do not need to be continuous in time ($\mathbb{C}^2$ continuity), something that intuitively should happen in real object motions. A simple example is shown in Fig. \ref{fig:fig1}. As long as the time between consecutive estimations is sufficiently small, discrete-time estimations might be accurate. However, this may not happen in real scenarios (e.g. due to temporal occlusions, low frame rates or fast dynamics).

In this work we address this by fitting the dynamic object trajectories to cubic cumulative B-Spline curves \cite{kim1995general}, a type of curve that in our proposal is defined by a series of $SE(3)$ control points distributed over time, which are interpolated to compute continuous poses. Such curve, has been previously used to estimate sensor ego-motion \cite{lovegrove2013spline}, but its application to 6-DoF object tracking remains unexplored. Moreover, we show for the first time the $SE(3)$ Jacobians of the pose with respect to the control points, thus significantly reducing its associated computational cost. In summary, our contributions are: 1) A RGB-D system able to estimate the 6-DoF trajectories of objects present in a scene, their related angular and linear velocities and accelerations, presenting all of them continuity with respect to time, and 2) the analytical derivation of the Jacobians of the interpolated pose with respect to the $SE(3)$ control points of a cubic B-Spline curve, applicable to any work that uses this type of curve.
    

\section{RELATED WORK}

\subsection{Object motion estimation}
One of the first works that aimed to estimate the motion of dynamic entities in the scene was \cite{tomasi1992shape}, in which the technique of \textit{Factorization} was introduced. To this end, several assumptions were made. Only one object was expected to be present and also an orthographic camera model was used, thus simplifying the computations at the expense of not considering the perspective projection of a real camera \cite{Hartley:2003:MVG:861369}. Follow-up works tackled these limitations. In \cite{sturm1996factorization} the perspective camera model was introduced, and \cite{han2004reconstruction, zappella2013joint} incorporated the motion estimation of multiple objects. Both aspects were addressed in \cite{sabzevari2014monocular}. However, the major part of these methods share some limitations: the difficulty of making them work sequentially, the need for specific motion assumptions, or high computational costs \cite{saputra2018visual}.


In the field of \textit{SLAM} the first work that incorporated the motion estimation of objects in its pipeline was \cite{wang2003online} (extended in \cite{wang2007simultaneous}), in which a Bayesian framework was proposed. Their results showed improvements with respect to only estimating the localization of the sensor. The same conclusion is reached in \cite{bibby2007simultaneous}, in which the 3D localization of the dynamic points are estimated. These works, as well as more recent ones \cite{bibby2010hybrid, kundu2011realtime, li2018stereo, yang2019cubeslam} are focused on objects whose movement is constrained to a plane, making them appropriate to environments like autonomous driving.

More recently, several works aim to estimate free 6-DoF motions of objects present in the scene. \cite{zhang2020vdo, bescos2021dynaslam} propose to first detect dynamic objects by using deep learning image segmentation techniques \cite{he2017mask}, to subsequently jointly estimate the object motions and the ego-motion of the sensor. In \cite{huang2020clustervo} a multi-level probabilistic association and a Conditional Random Field are added to ensure a correct data association between 3D points. Similarly \cite{judd2018multimotion, judd2021multimotion} propose a clustering approach based on the 3D motion of points to associate them to each object. In our work we follow a similar deep learning-based strategy using SiamMask \cite{wang2019fast}.

\subsection{Continuous-time tracking}

For a general view of interpolation methods for tracking, we refer the reader to the thorough survey of Haarbach et al. \cite{haarbach2018survey}. In this work, we focus on cumulative B-Splines, which were originally defined in \cite{kim1995general} for the graphics field. As highlighted in \cite{patron2015spline}, B-Splines present interesting properties for robotics/vision tasks, 
which were leveraged for the first time in \cite{lovegrove2013spline} to estimate the trajectory of a rolling shutter camera for calibration and visual-inertial SLAM. Later works extended their use for ego-motion estimation with other sensors: \cite{kerl1dense} for RGB-D, \cite{mueggler2018continuous} for event cameras, \cite{droeschel2018efficient} for 3D laser-range scanners and more recently in \cite{yang2021asynchronous} for a multi-camera set-up. We also find applications in SfM \cite{ovren2019trajectory}\red{, in which is also shown that $SE(3)$ B-Splines are preferred over the split $SO(3)\times\mathbb{R}^3$ representation when force and torque are related, which holds in general for rigid object motions \cite{lynch2017modern}.}

In all previous works, the Jacobians were computed either with automatic differentiation (mainly the implementation of \cite{ceres-solver}) or with numerical differentiation. As noted by some authors \cite{mueggler2018continuous}, the analytical derivation of the Jacobians is a critical step to drastically reduce the execution time. Recently in \cite{sommer2020efficient} the Jacobians for the $SO(3)$ cumulative B-Splines were derived. In this work, we derive them for $SE(3)$. 

Related to our work, \cite{bibby2010hybrid} implements a continuous-time estimation of the object motions by using splines. However, a planar motion assumption ($SE(2)$) is made. To the best of our knowledge, our work is the first to apply continuous-time object tracking for 6-DoF. 

\section{BACKGROUND}

\subsection{SE(3) Lie Group}
A reference frame $\{o\}$, that is attached to an object, can be expressed with respect to a global reference frame $\{w\}$ with a \textit{transformation matrix} $\bT_{wo}\in SE(3)$:
\begin{equation}
    \bT_{wo} = \begin{bmatrix}
    \bR_{wo} & \bt_{wo}\\
    \bzero^\top & 1
    \end{bmatrix},
\end{equation}
where $\bR_{wo}\in SO(3)$ and $\bt\in\mathbb{R}^3$ encode respectively, the orientation and translation of $\{o\}$ with respect to $\{w\}$. 
$SE(3)$ is both a group and a smooth manifold, 
implying that at each point, $\bT\in SE(3)$, exists a unique tangent space called Lie Algebra or $se(3)$, which can be defined locally at $\bT$, and at the identity $\bI$ \cite{sola2018micro}. An element $\btauhat\in se(3)$ has the form:
\begin{equation}
    \btauhat=\begin{bmatrix}
    \mathbf{v}\\
    \bomega
    \end{bmatrix}^{\wedge} = \begin{bmatrix}
    \bomegahat & \mathbf{v}\\
    \mathbf{0}^\top & 0 
    \end{bmatrix},
\end{equation}
where $\bv\in\mathbb{R}^3$ and $\bomegahat$ is the anti-symmetric matrix related to $\bomega\in\mathbb{R}^3$. $(\cdot)^\wedge$ is the \textit{hat} operator, used for a convenient vectorization. An element $\btauhat\in se(3)$ is mapped to $SE(3)$ and vice versa via the \textit{exponential} and \textit{logarithm} mappings:
\begin{alignat}{7}
    \Exp & \quad:\quad  \Rsix && \mapsto SE(3)\quad&;&\quad \btau &&\mapsto \bT &&= \Exp(\btau),\\
    \Log & \quad:\quad  SE(3) && \mapsto \Rsix\quad&;&\quad \bT &&\mapsto \btau &&= \Log(\bT),
\end{alignat}
where we have used the capitalized notation of \cite{sola2018micro}.

\begin{figure*}[t]
    \centering
    \begin{subfigure}[t]{0.30\linewidth}
      \centering
    \includegraphics[width=1.0\textwidth]{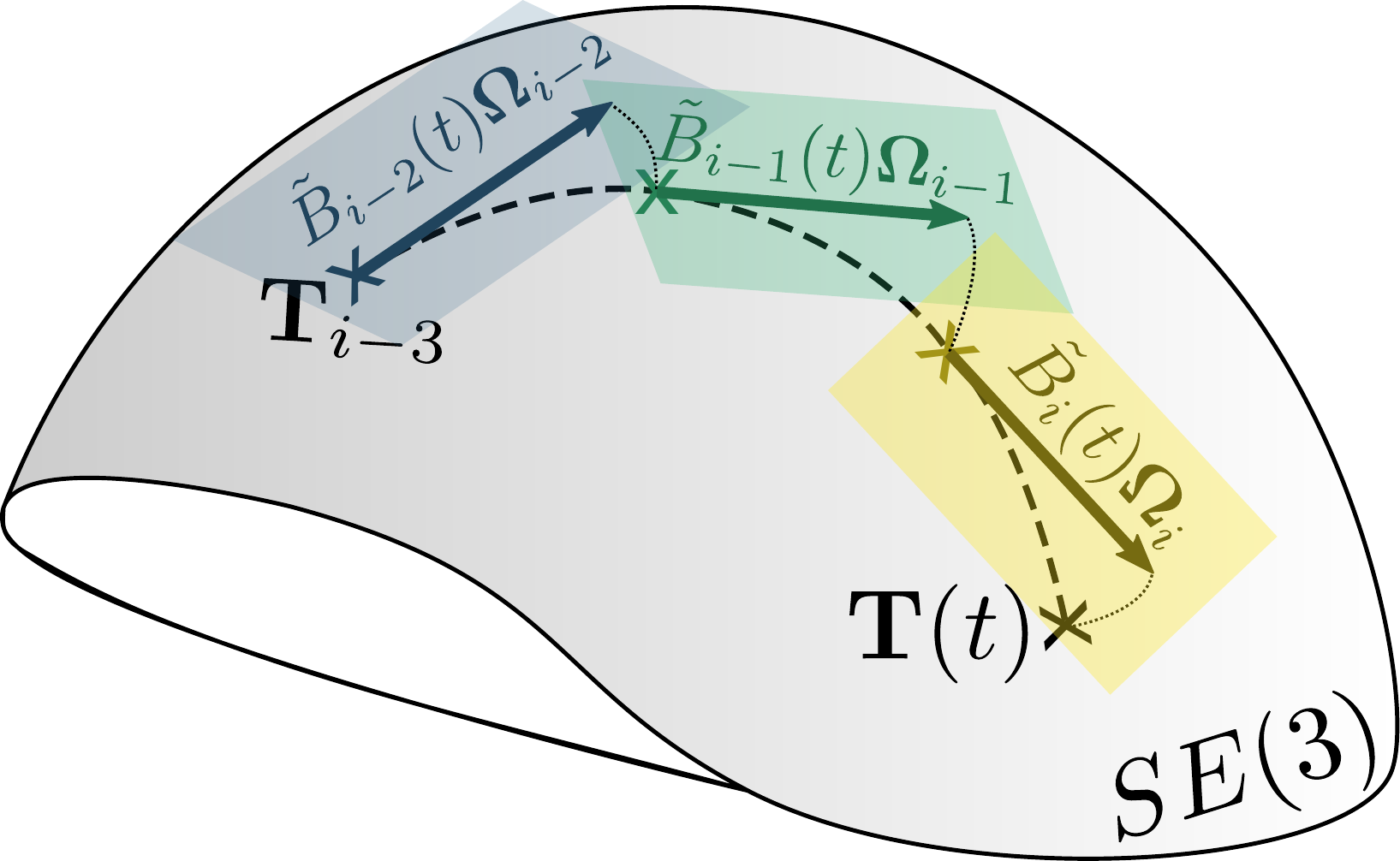}
      \caption{}
      \label{fig:fig2}
    \end{subfigure}
    \hfill
    \begin{subfigure}[t]{0.30\linewidth}
      \centering
    \includegraphics[width=1.0\textwidth]{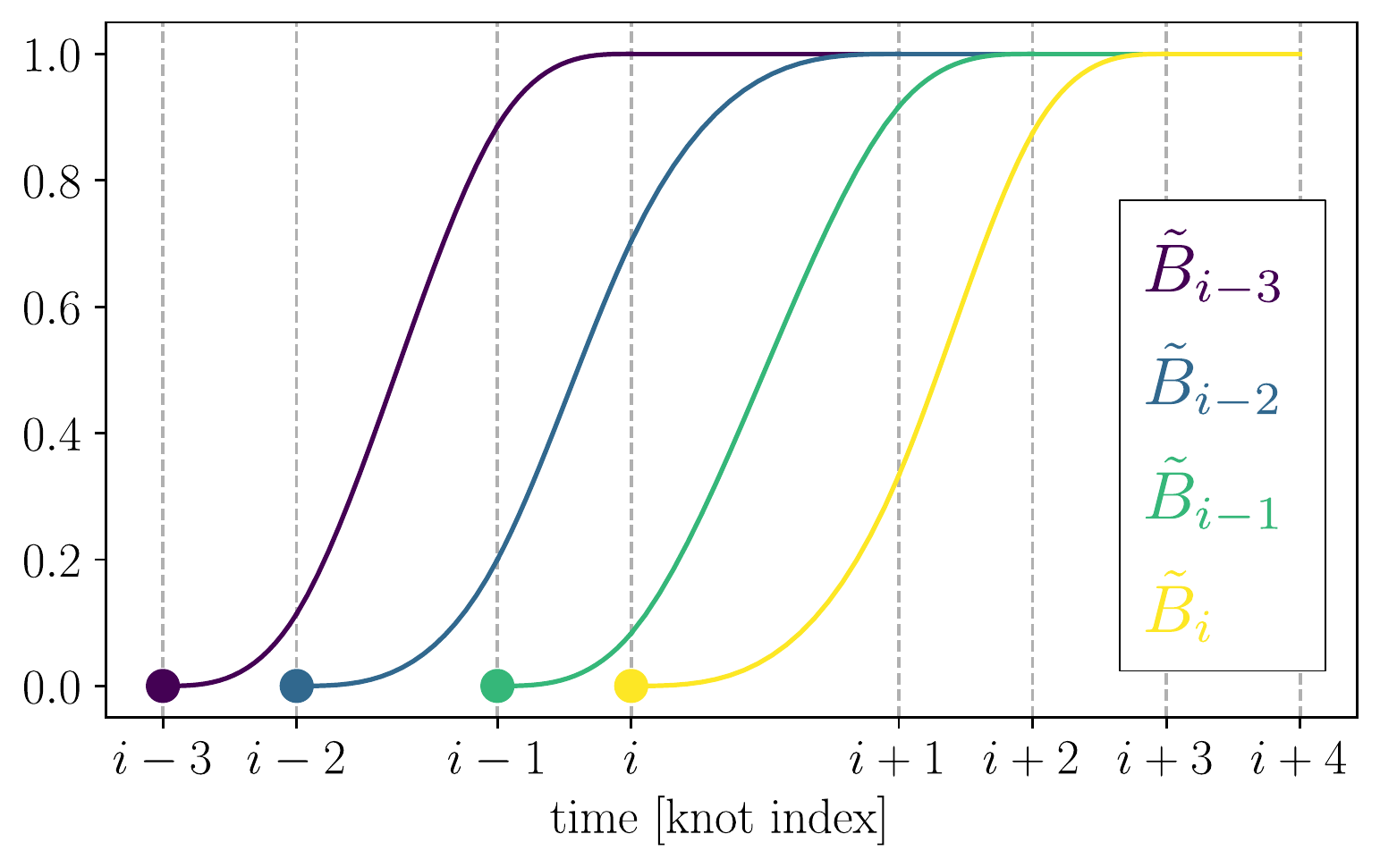}
      \caption{}
      \label{subfig:cumul_basis}
    \end{subfigure}
    \hfill
    \begin{subfigure}[t]{0.36\linewidth}
      \centering
      \includegraphics[width=1.0\textwidth]{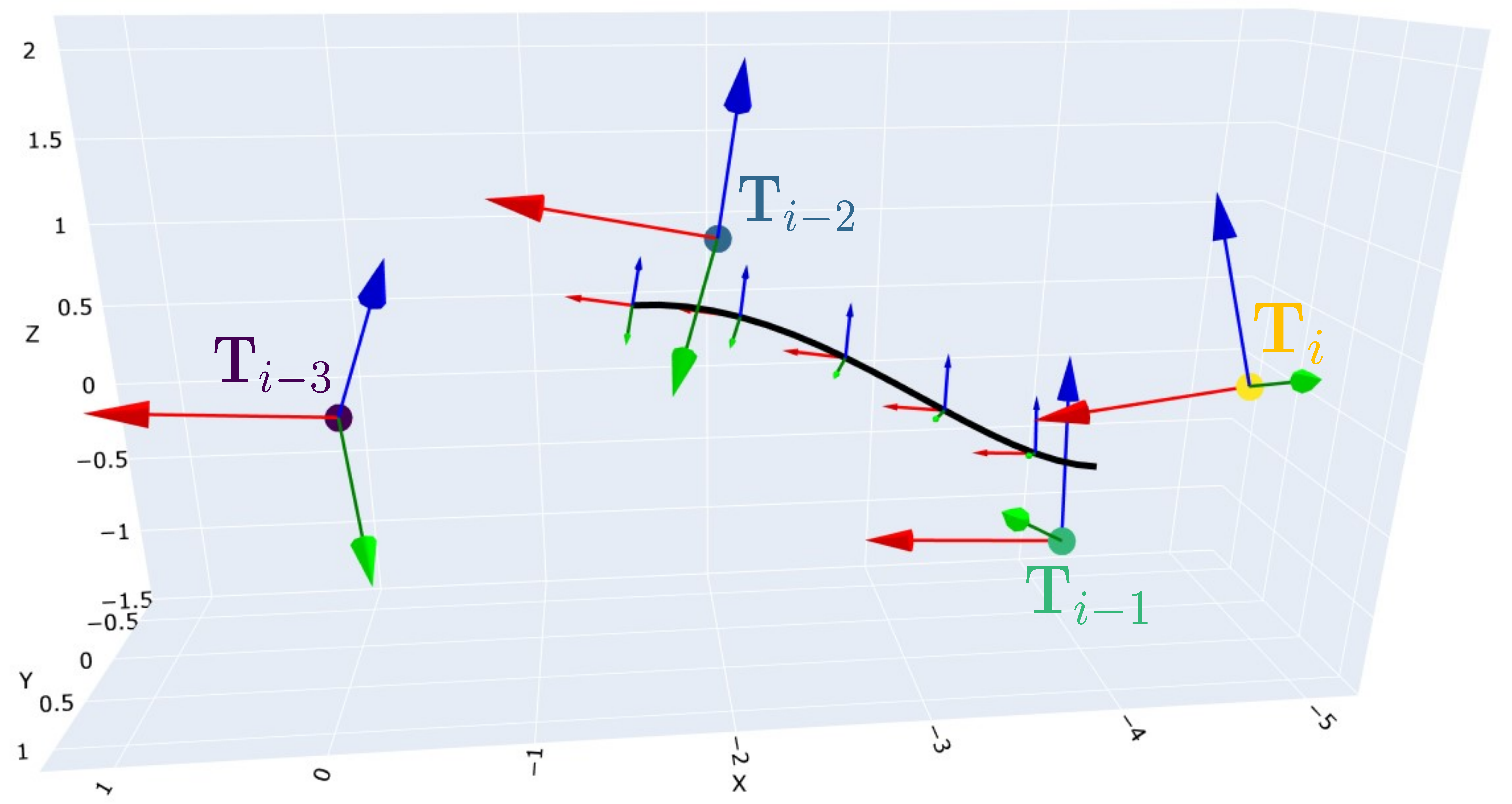}
      \caption{}
      \label{subfig:interp}
    \end{subfigure}
    \caption{\red{Interpolation of a pose $\bT(t)\in SE(3)$ with $t\in[t_i, t_{i+1})$ and control points $\{\bT_{i-3}\,..\,\bT_{i}\}$. (a) Visualization of Eq. \ref{eq:Tinter_cumul2} in the $SE(3)$ manifold. The successive compositions, $\Exp(\tilde{B}_{j}(t)\bOmega_{j})$ with $j\in\{i-3\,..\,i\}$, are parameterized over the successive local tangent spaces. (b) Cumulative basis functions with influence at $t\in[t_i,t_{i+1})$. (c) 4 exemplar control points (bigger reference frames) and resultant interpolation at this time span. For clarity, only a few interpolated frames are shown.}}
    \label{fig:fig3}
    \vspace{-0.5cm}
\end{figure*}

This way, we can compose a transformation matrix $\bT_{wo}$ with another parameterized in the local tangent space: $\bT_{wo}\,\Exp(\btau_o)$, or in the tangent space defined at the identity: $\Exp(\btau_w)\bT_{wo}$. The equivalence between the two is given by the \textit{Adjoint matrix} $\text{Ad}_{\bT_{wo}}\in\mathbb{R}^{6\times6}$: $\btau_w = \text{Ad}_{\bT_{wo}}\btau_o$. As a result (used in Sec. \ref{subsec:jacder}), we have that:
\begin{equation}\label{eq:adj}
    \Exp(\btau_w)\bT_{wo} = \bT_{wo}\Exp(\text{Ad}_{\bT_{wo}^{-1}}\btau_w) = \bT_{wo}\Exp(\btau_o).
\end{equation}

The elements of $se(3)$ can also be related to the kinematics of the coordinate system $\{o\}$ \cite{lynch2017modern}. Using Newton's notation for differentiation with respect to time: 
\begin{equation}
\btauhat_o=
\begin{bmatrix}
    \bv_o\\
    \bomega_o
\end{bmatrix}^\wedge=
\bT_{wo}^{-1}\dot{\bT}_{wo}=
\begin{bmatrix}
    \bR_{wo}^\top \dot{\bR}_{wo} & \bR_{wo}^\top\dot{\bt}_{wo}\\
    \bzero^\top & 0 
\end{bmatrix},  \label{eq:kin}
\end{equation}
contains the linear $\bv_o$ and angular $\bomega_o$ velocities of $\{o\}$ expressed in a coordinate system that is fixed and instantaneously coincident with $\{o\}$. The linear and angular accelerations are obtained time-differentiating again Eq. \ref{eq:kin}. These quantities can be transformed to $\{w\}$ via $\bR_{wo}$.

Because of its importance in the Jacobian derivations (Sec. \ref{subsec:jacder}), we introduce the \textit{left Jacobian} $\bJl$ of $SE(3)$ \cite{sola2018micro}:
\begin{equation}
    \bJ_l(\btau)=\left.\frac{\partial\Log(\Exp(\btau+\delta\btau) \Exp(\btau)^{-1}\redd{)}}
    {\partial\delta\btau}\right|_{\delta\btau=\bzero}.
\end{equation}
It maps variations of $\btau$ to variations expressed in the tangent space at the identity and composed with the current pose. Two results that will be used are that, for small $\bxi\in\mathbb{R}^6$:
\begin{align}
    \Exp(\btau+\bxi) &\approx \Exp(\bJl(\btau)\bxi)\Exp(\btau), \label{eq:Jl1}\\
    \Log(\Exp(\bxi)\Exp(\btau)) &\approx \btau + \bJ_l^{-1}(\btau)\bxi. \label{eq:Jl2}
\end{align}

Closed-form expressions for: $\Exp(\btau), \Log(\bT), \text{Ad}_\bT$ and $\bJ_l(\btau)$, with $\btauhat\in se(3), \bT\in SE(3)$, can be found in \cite{barfoot2017state}.

\subsection{Cumulative B-Splines}\label{sec:Bsplines}

A pose $\bT(t)\in SE(3)$ at time $t$, interpolated with a cumulative B-Spline \red{of $n+1$ control points} is given by \cite{lovegrove2013spline}:
\begin{equation}\label{eq:Tinter_cumul}
    \mathbf{T}(t) = \Exp(\tilde{B}_{0,k}(t)\Log(\bT_0))\prod_{i=1}^{n}\Exp(\tilde{B}_{i,k}(t)\bOmega_i),
\end{equation}
where $\tilde{B}_{i,k}(t)$ is the $i-th$ \red{scalar} \textit{cumulative basis} function, a $\mathbb{C}^{k-2}$ continuous polynomial of degree $k-1$. Each one is related to a time stamp $t_i$ (\textit{knot}), satisfying:
\begin{equation}\label{eq:cumul_basis}
    \tilde{B}_{i,k}(t) = 
    \begin{cases}
        0 & \text{if } t \leq t_{i}\\
        \sum_{j=i}^{i+k}B_{j,k}(t) & \text{if } t\in(t_i,t_{i+k-1})\\
        1 & \text{if } t \geq t_{i+k-1}
    \end{cases}
\end{equation}
where the terms $B_{j,k}(t)$ are the standard B-Spline basis functions obtained with the de Boor-Cox formula \cite{cox1972numerical}. Eq. \ref{eq:cumul_basis} conditions are visualized in Fig. \ref{subfig:cumul_basis}. 

For $i>0$ each cumulative basis weighs the relative difference between the \textit{control points} $\bT_{i-1}, \bT_i\in SE(3)$ i.e. $\bOmega_i=\red{\text{Log}(}\bT_{i-1}^{-1}\,\bT_i\red{)}$. These control points are the variables to be estimated. Since we are interested in a curve with $\mathbb{C}^2$ continuity, $k=4$ (\textit{cubic} cumulative B-Spline) is chosen, thereby $\tilde{B}_{i,4}(t)=1$ if $t\geq t_{i+3}$ (Eq. \ref{eq:cumul_basis}). Using this fact, Eq. \ref{eq:Tinter_cumul} for $t\in[t_i,t_{i+1})$ can be simplified to\footnote{From \red{now on}, we drop out the subscript $k=4$ 
to avoid clutter.}:
\begin{equation}\label{eq:Tinter_cumul2}
    \mathbf{T}(t) = \bT_{i-3}\prod_{j=1}^{3}\Exp(\tilde{B}_{i-3+j}(t)\bOmega_{i-3+j}).
\end{equation}
Time derivatives $\dot{\bT}$ and $\ddot{\bT}$, are given in \cite{lovegrove2013spline, sommer2020efficient}.
Fig. \ref{fig:fig2} shows a visual conceptualization of Eq. \ref{eq:Tinter_cumul2} in the manifold of $SE(3)$. Additionally, in Fig. \ref{subfig:interp} we show 4 exemplar control points and the resultant interpolation using the cumulative basis functions of Fig. \ref{subfig:cumul_basis}.

In our implementation, to compute the value of a cumulative basis function at time $t\in[t_i,t_{i+1})$, we combine the cumulative definition of Eq. \ref{eq:cumul_basis} with the matrix representation of the standard basis functions derived in \cite{qin2000general}:
\begin{align}
    \tilde{\mathbf{B}} &= \begin{bmatrix}
    \tilde{B}_{i-3}(t) & \tilde{B}_{i-2}(t) & \tilde{B}_{i-1}(t) & \tilde{B}_{i}(t)
    \end{bmatrix} = \mathbf{u}^\top \tilde{\mathbf{M}}, \nonumber\\
    \tilde{\mathbf{M}} &= 
    \begin{bmatrix}
        1 & 1-m_{00} & m_{02} & 0\\
        0 & 3m_{00}  & m_{12} & 0\\
        0 & -3m_{00} & m_{22} & 0\\
        0 & m_{00}   & m_{32} + m_{33} & m_{33}
    \end{bmatrix}, \label{eq:quin_cumul}
\end{align}
where $\mathbf{u} = \begin{bmatrix} 1 & u & u^2 & u^3 \end{bmatrix}^\top$, with $u=\frac{t-t_i}{t_{i+1}-t_i}\in[0,1)$. The terms $m_{ij}$ correspond to the ones defined in \cite[Sec.~3.2]{qin2000general}. $\tilde{\mathbf{M}}$ is the \red{cumulative reformulation without assuming the common constraint \cite{lovegrove2013spline, mueggler2018continuous} of constant time intervals between control points. This can benefit applications that require more flexibility in their placement}.

\section{CONTINUOUS-TIME OBJECT TRACKING}

At each time stamp $t$, our approach receives as inputs a RGB-D image,  
its estimated pose $\bT_{wc}\in SE(3)$
, and instance segmentation masks without inter-image associations (we used COLMAP \cite{schonberger2016structure} and SiamMask \cite{wang2019fast} in our experiments). The outputs are the continuous-time trajectories $\bT_{wo}(t)$ of each observed object
, parameterized with the control points of a cumulative B-Spline (Sec. \ref{sec:Bsplines}). Since our formulation is independent for each object, we use one common subscript. An additional optional output is the set of optimized 3D objects' points in the object frame $\{o\}$. Two main blocks conform the proposal: the front-end (Sec. \ref{subsec:frontend}) and the optimization back-end (Sec. \ref{subsec:optim}). The front-end is in charge of 1) object correspondences between frames, 2) initialization of new trajectories, and 3) providing enough object feature tracks to optimize its trajectory. Special care is taken in the filtering of outliers. The back-end receives this information and optimizes the set of control points, $\mathcal{T}$, of each object's continuous-time trajectory and optionally a set of sparse objects' points $\mathcal{P}$. To this end, a robustified Gauss-Newton 
algorithm is used.

\subsection{Optimization}\label{subsec:optim}
To avoid an unconstrained increase in computational cost, we adopt a sliding window optimization. The set of 3D object point observations $\bp_c\in\mathbb{R}^3$ expressed in $\{c\}$ within this temporal window is denoted by $\mathcal{Z}$. We adopt an object-centric parameterization \cite{bescos2021dynaslam}, i.e. we relate each observation $\bp_c\in\mathcal{Z}$ to its correspondent point $\bp_o\in\mathcal{P}$ in the object frame $\{o\}$. If an object point $\bp_o$ is observed in different images, multiple observations in $\mathcal{Z}$ are related to it. 

We define the estimated 3D location error, $\mathbf{e}_{\bp_c}\in\mathcal{E}$ (with $\mathcal{E}$ as the set of errors within the temporal window)
as:
\begin{equation}\label{eq:error}
    \mathbf{e}_{\bp_c} = \bp_c - \proj\left(\bT_{wc}^{-1}\,\bT_{wo}(t)\,\tilde{\bp}_o\right),
\end{equation}
\begin{wrapfigure}{r}{2.8cm}
\vspace{-0.5cm}
\includegraphics[width=2.8cm]{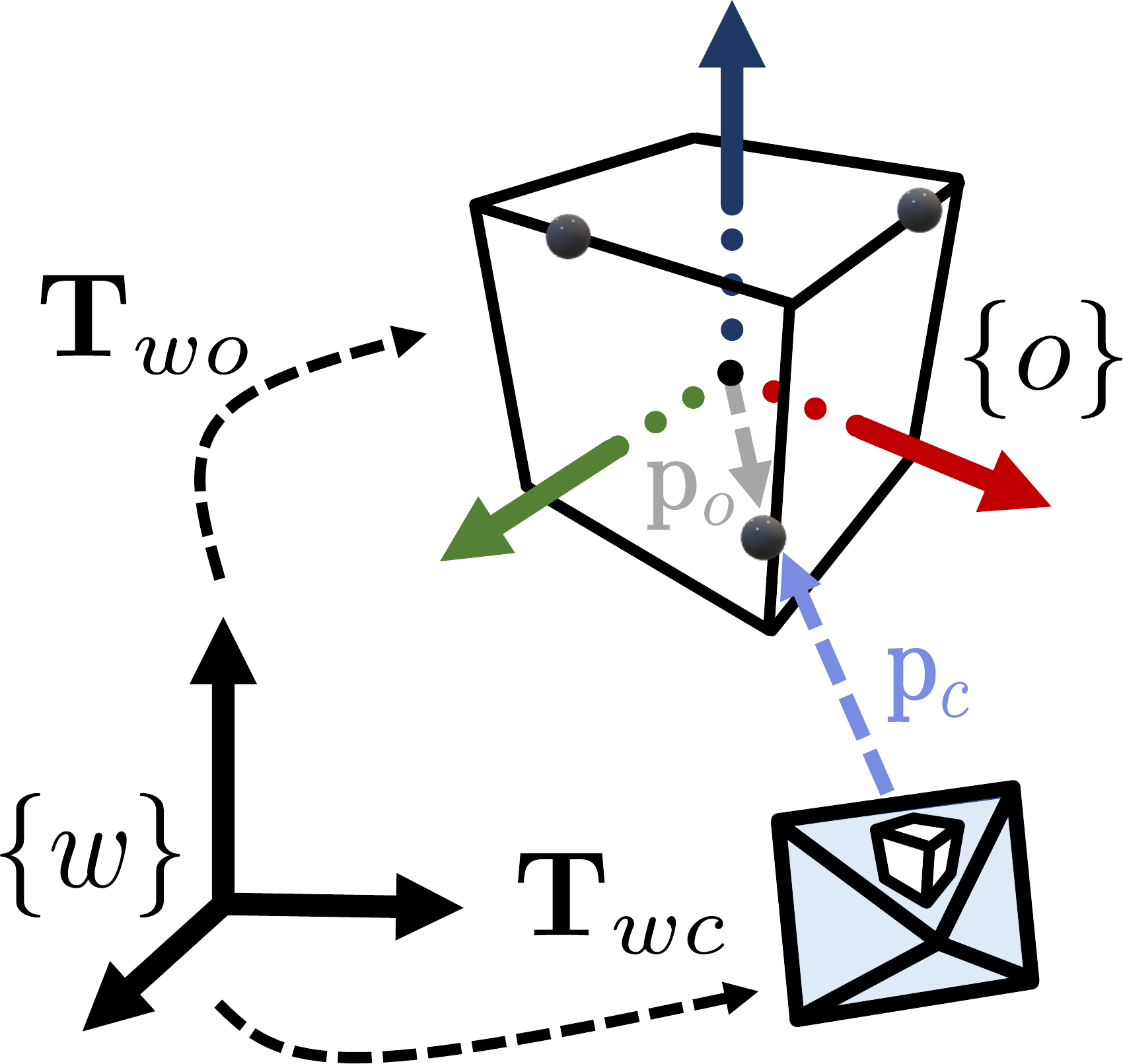}
\caption{Notation.}\label{wrap-fig:1}
\vspace{-0.3cm}
\end{wrapfigure}
where $\tilde{\bp}_o\in\mathbb{P}^3$ is the homogeneous representation of $\bp_o$, $\bT_{wo}(t)$ is the interpolated pose (with Eq. \ref{eq:Tinter_cumul2}) at the time stamp $t$ at which $\bp_c$ was observed from a camera with estimated pose $\bT_{wc}$. $\text{proj}: \mathbb{P}^3\mapsto\mathbb{R}^3$ performs the homogeneous to Cartesian coordinates mapping.

We denote as $\mathcal{X}$ the set of parameters that influences $\mathcal{E}$. If only the control points of the trajectories are optimized, then $\mathcal{X}\equiv\mathcal{T}$. If the objects' points are also optimized then $\mathcal{X}\equiv\{\mathcal{T},\mathcal{P}\}$. We refer to these optimizations as \textit{Spline BA} and \textit{Local BA}. Then we seek to minimize $E(\mathcal{X})$:
\begin{equation}\label{eq:BA3d}
    E(\mathcal{X}) =
    \frac{1}{2}
    \sum_{\substack{\be_{\bp_c}\in\mathcal{E}
    }}
    \rho\left(
    \be_{\bp_c}^\top \bSigma^{-1}_{\bp_c} \be_{\bp_c}
    \right).
\end{equation}
Assuming observations to be independent and perturbed with 
zero-mean Gaussian noise with covariance 
$\bSigma_{\bp_c}$ \red{(without prior knowledge, we set it to the identity)}, minimizing Eq. \ref{eq:BA3d} leads to maximizing the likelihood $\mathcal{L}(\mathcal{X}|\mathcal{Z})$ \cite{dellaert2017factor}. The Huber loss $\rho(\cdot)$ \cite{huber2004robust} is used to reduce the influence of outliers.

To minimize Eq. \ref{eq:BA3d} we iteratively perform updates $\Delta\bx$ on the parameters $\bx$ (vectorized $\mathcal{X}$) by solving the normal equations: $\bH\Delta\bx = -\mathbf{g}$, where $\bH$ and $\bg$ are the (approximate) Hessian and gradient of $\be$ (vectorized $\mathcal{E}$) with respect to $\bx$. For each observation they are computed as:
\begin{equation}
    \bH_{\be_{\bp_c}} = \rho'\bJ_{\be_{\bp_c}}^\top\bSigma^{-1}_{\bp_c}\bJ_{\be_{\bp_c}},
    \quad
    \bg_{\be_{\bp_c}} = \rho' \be_{\bp_c}^\top \bSigma^{-1}_{\bp_c} \bJ_{\be_{\bp_c}},
\end{equation}
where $\rho'$ is the Huber loss derivative at $\be_{\bp_c}^\top \bSigma^{-1}_{\bp_c} \be_{\bp_c}$, and $\bJ_{\be_{\bp_c}} = \partial\be_{\bp_c}/\partial\bx$ is the Jacobian of the observation error.

\subsection{Jacobians Derivation}\label{subsec:jacder}

To compute each $\bJ_{\be_{\bp_c}}$ we need to differentiate $\be_{\bp_c}$ with respect to the control points. To this end, as it is common when dealing with $SE(3)$ poses \cite{barfoot2017state}, we parameterize each control point $\bT_i$ with a perturbation $\bxi_i^\wedge\in se(3)$, so that:
\begin{equation}
    \left.\frac{\partial\be_{\bp_c}(\Exp(\bxi_i)\bT_i)}{\partial \bxi_i}\right|_{\bxi_i=\bzero}
    =\frac{\partial\be_{\bp_c}(\Exp(\bxi_i)\bT_i)}{\partial \bxi_i}(\bzero),
\end{equation}
is used to update $\bT_i\leftarrow\Exp(\bxi_i)\bT_i$ with the value of $\bxi_i\in\mathcal{X}$ computed at each iteration of the optimization process.

It is of special interest the derivative of an interpolated pose $\bT(t)$ (Eq. \ref{eq:Tinter_cumul2}) w.r.t. the control points as it has not been addressed yet in the literature and would imply significant computational savings \cite{mueggler2018continuous}. \redd{In this section we derive them for its $se(3)$ form and for its 12-dimensional vectorized form of the $SE(3)$ object}\footnote{\redd{Both forms can fit a wide range of cost functions via the chain rule. In Sec. \ref{sec:exp_timings} we show their benefits in our continuous-time tracking problem.}}.
The notation used is the following (for convenience, we particularize Eq. \ref{eq:Tinter_cumul2} with $i=3$):
\begin{alignat}{2}
    &\bT(t) &&= \Exp(\ba_0)\bT_0\bA_1(t)\bA_2(t)\bA_3(t),\label{eq:app1}\\
    &\bA_j(t) &&= \Exp(\ba_j(t)),\label{eq:app2}\\
    &\ba_j(t) &&=\tilde{B}_j(t)\bOmega_j,\label{eq:app2b}\\
    &\ba_0 &&= \bxi_0,\label{eq:a0_def}\\
    &\bOmega_j &&= \Log\left((\Exp(\bxi_{j-1})\bT_{j-1})^{-1}\Exp({\bxi_j})\bT_j)\right),\label{eq:app3}
\end{alignat}
with $j\in\{1,2,3\}$ and the perturbations evaluated at $\bzero$. 

\redd{Focusing first on the 12-d vectorized form, $\bT_{\vect}$ (see Eq. \ref{eq:vect}), and} using the multi-variable chain rule, we have:
\begin{alignat}{3}
    &\bxi_{j\in\{0,1,2\}} &&: 
    \frac{\pt\bT_{\vect}}{\pt\bxi_j}(\bzero) =
    \frac{\pt\bT_{\vect}}{\pt \ba_j}\frac{\pt\ba_j}{\pt\bxi_j}(\bzero) + 
    \frac{\pt\bT_{\vect}}{\pt \ba_{j+1}}\frac{\pt\ba_{j+1}}{\pt\bxi_j}(\bzero),\nonumber\\
    &\bxi_3 &&:
    \frac{\pt\bT_{\vect}}{\pt\bxi_3}(\bzero) = 
    \frac{\pt\bT_{\vect}}{\pt \ba_3}\frac{\pt\ba_3}{\pt\bxi_3}(\bzero),\label{eq:chain}
\end{alignat}
where $\bT_{\vect}$ vectorizes $\bT$ to represent it as a 1D vector:
\begin{equation}\label{eq:vect}
    \bT_{\vect} = \begin{bmatrix}
    (\bR^{c1})^\top & (\bR^{c2})^\top & (\bR^{c3})^\top & \bt^\top
    \end{bmatrix}^\top,
\end{equation}
with $\bR^{ci}$ as the $i-th$ column of $\bR$ (rotation in $\bT$). The last row is ignored since it would add meaningless computations.
With these considerations \red{(and definitions of Eqs. \ref{eq:extra1}-\ref{eq:extra2})}:
\begin{align}
    \frac{\pt\bT_{\vect}}{\pt\ba_j} 
    &= \frac{\pt(\bP_j\Exp(\ba_j +\btau_j)\bN_j)_{\vect}}{\pt\btau_j}(\bzero),\\
    &\overset{\ref{eq:Jl1}}{=} \frac{\pt}{\pt\btau_j} (\bP_j
    \underbrace{\Exp(\bJ_{l}(\ba_j)\btau_j)}_{\bC(\btau_j)}
    \underbrace{\Exp(\ba_j)\bN_j}_{\bN_j'})_{\vect}(\bzero),\label{eq:usingJl}\\
    &= \frac{\pt(\bP_j\bC(\btau_j)\bN_j')_{\vect}}{\pt \bC(\btau_j)_{\vect}}
    \frac{\pt\bC(\btau_j)_{\vect}}{\pt\bJ_{l}(\ba_j)\btau_j}
    \frac{\pt \bJ_{l}(\ba_j)\btau_j}{\pt \btau_j}\bigg|_{\btau_j=\bzero},\label{eq:2nd_cr} \raisetag{25pt}
\end{align}
with all the derivatives of Eq. \ref{eq:2nd_cr} evaluated at $\btau_j=\bzero$, and:
\begin{alignat}{3}
    &j=0 &&\quad\to\quad \bP_0=\bI_{4\times4}, \quad &&\bN_0'= \bT,\label{eq:extra1}\\
    &j=1 &&\quad\to\quad \bP_1=\bT_0, \quad &&\bN_1'= \bA_1\bA_2\bA_3,\\
    &j=2 &&\quad\to\quad \bP_2=\bT_0\bA_1, \quad &&\bN_2'= \bA_2\bA_3,\\
    &j=3 &&\quad\to\quad \bP_3=\bT_0\bA_1\bA_2, \quad &&\bN_3'= \bA_3.\label{eq:extra2}
\end{alignat}

The right-most term of Eq. \ref{eq:2nd_cr} is straightforward:
\begin{equation}
    \frac{\pt \bJ_{l}(\ba_j)\btau_j}{\pt \btau_j}(\bzero) = \bJ_{l}(\ba_j),\label{eq:app4}
\end{equation}
Note that $\bJ_l(\ba_0)|_{\bxi_0=\bzero}=\bI_{6\times6}$.
For the left-most term of Eq. \ref{eq:2nd_cr}, denoting the Kronecker product as $\otimes$, and the rotation matrix of $\bP_j$ as $\bR_{\bP_j}$, from \cite[Eq.~11-12]{blanco2010tutorial}:
\begin{equation}
    \frac{\pt(\bP_j\bC(\btau_j)\bN_j')_{\vect}}{\pt \bC(\btau_j)_{\vect}}(\bzero) = (\bN_j')^\top \otimes \bR_{\bP_j},
\end{equation}

The middle term of Eq. \ref{eq:2nd_cr} is given by the generators of $SE(3)$, $\{\bG_i\}_{i=1}^{6}$ \cite[Eq.~A.1]{strasdat2012local}, since they map, at the identity, infinitesimal variations in the dimensions of an element $\btau^\wedge\in se(3)$ to variations in $SE(3)$:
\begin{equation}\label{eq:gen}
    \frac{\pt\Exp(\bJ_{l}(\ba_j)\btau_j)_{\vect}}{\pt\bJ_{l}(\ba_j)\btau_j}(\bzero) =
    \begin{bmatrix}
    (\bG_1)_{\vect} & \ldots & (\bG_6)_{\vect}
    \end{bmatrix} = \bG,
\end{equation}
where $(\bG_i)_{\vect}$ indicates (with slight abuse of notation) the same vectorization as in Eq. \ref{eq:vect}, i.e., $\bG$ is a $12\times6$ matrix.

It only remains to derive $\pt\ba_{j}/\pt\bxi_j$ and $\pt\ba_{j+1}/\pt\bxi_j$. A useful observation is that, for $j\in\{1,2,3\}$:
\begin{equation}
    (\Exp(\bxi_{j-1})\bT_{j-1})^{-1} = \bT_{j-1}^{-1}\Exp(-\bxi_{j-1}),
\end{equation}
so we only need to derive $(\pt\ba_j/\pt \bxi_{j})|_{\bxi_j=\bzero}$, since:
\begin{equation}
    \frac{\pt\bOmega_j}{\pt \bxi_{j-1}} = -\frac{\pt\bOmega_j}{\pt \bxi_{j}},\qquad
    \frac{\pt\ba_j}{\pt \bxi_{j-1}} = -\frac{\pt\ba_j}{\pt \bxi_{j}},
\end{equation}
which can be obtained as follows:
\begin{align}
    \frac{\pt\ba_j}{\pt \bxi_{j}}(\bzero)
    &= \frac{\pt\tilde{B}_j(t)
    \Log(\bT_{j-1}^{-1}\Exp({\bxi_j})\bT_j)}{\pt \bxi_{j}}(\bzero),\\
    &\overset{\ref{eq:adj}}{=} \frac{\pt\tilde{B}_j(t)  \Log(\Exp(\text{Ad}_{\bT_{j-1}^{-1}}\bxi_j)\bT_{j-1}^{-1}\bT_j
    )}{\pt \bxi_{j}} (\bzero),\\
    &\overset{\ref{eq:Jl2}}{=} \frac{\pt\tilde{B}_j(t)\bJ_l^{-1}(\Log(\bT_{j-1}^{-1}\bT_j))\text{Ad}_{\bT_{j-1}^{-1}}\bxi_j}{\pt \bxi_{j}}
    (\bzero),\label{eq:usingJl2}\\
    &= \tilde{B}_j(t)\,\,\bJ_l^{-1}(\Log(\bT_{j-1}^{-1}\bT_j))\text{Ad}_{\bT_{j-1}^{-1}}.
\end{align}
Lastly, $(\pt\ba_0/\pt \bxi_{0})(\bzero) = \bI_{6\times6}$ (from its definition at Eq. \ref{eq:a0_def}). Note that both Eq. \ref{eq:usingJl} and \ref{eq:usingJl2} are exact since they are evaluated at $\bzero$. Hence, only the first-order term has influence. 

\red{
Now, we focus on the Jacobian of the minimal $se(3)$ representation, $\Log(\bT)$, w.r.t. the perturbations. Starting off Eq. \ref{eq:chain}, 
$\frac{\pt\bT_{\vect}}{\pt \ba_k}$
($k\in\{0\,..\,3\}$) is the only term that we need to change in favor of 
$\frac{\pt\Log(\bT)}{\pt \ba_k}$. For $\ba_\bzero$ it can be obtained as:
\begin{align}
    \frac{\pt\Log(\bT)}{\pt\ba_0} 
    &= \frac{\pt\Log(\Exp(\ba_0)\bT)}{\pt\ba_0}
    \overset{\ref{eq:Jl2}}{=} \bJ_l^{-1}(\Log(\bT)).
\end{align}
Since $\ba_0=\bxi_0$ evaluated at $\bzero$. Lastly, for $k=\{1,2,3\}$:
\begin{align}
    \frac{\pt\Log(\bT)}{\pt\ba_k} 
    &= \frac{\pt \Log(\bP_{k}\Exp(\ba_{k} +\btau_k)\bN_{k+1}')}{\pt \btau_k}(\bzero),\\
    &\overset{\ref{eq:adj},\ref{eq:Jl1}}{=} \frac{\pt\Log(\Exp(\text{Ad}_{\bP_k}\bJ_l(\ba_k)\btau_k)\bT)}{\pt \btau_k}(\bzero),\\
    &\overset{\ref{eq:Jl2}}{=}\bJ_l^{-1}(\Log(\bT))\text{Ad}_{\bP_k}\bJ_l(\ba_k),\label{eq:app5}
\end{align}
which is similar to \cite[Eq. 57]{sommer2020efficient} but without imposing the adjoint of SO(3). To sum up, the Jacobians of both representations are completely defined by (with $k\in\{0\,..\,3\}$):
\begin{alignat}{3}
    &\frac{\pt\ba_k}{\pt \bxi_{k}}(\bzero) && =
    \tilde{B}_k(t)\,\,\bJ_l^{-1}(\Log(\bT_{k-1}^{-1}\bT_k))\text{Ad}_{\bT_{k-1}^{-1}},\label{eq:sumup1}\\
    &\frac{\pt\bT_{\vect}}{\pt \ba_k} &&= 
    ((\bN_k')^\top \otimes \bR_{\bP_k})\,\bG\,\bJ_l(\ba_k),\label{eq:sumup2}\\
    & \frac{\pt\Log(\bT)}{\pt\ba_k}&&=\bJ_l^{-1}(\Log(\bT))\text{Ad}_{\bP_k}\bJ_l(\ba_k).\label{eq:extra4}
\end{alignat}
}
\redd{For completeness (although not used in our tracking method), we show how to extend them to higher degrees and the analytic Jacobians of the velocity in $SE(3)$ as an appendix\footnote{\redd{See Appendices I, II of: \url{https://arxiv.org/abs/2201.10602}}}.}

\subsection{Implementation details related to the control points}

\begin{figure}[tb]
    \centering 
    \includegraphics[width=0.49\textwidth]{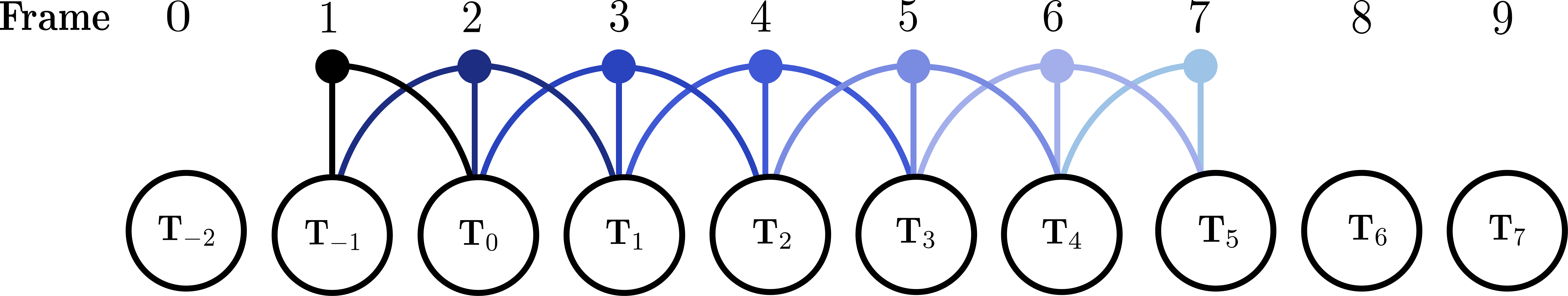}
    \caption{Simple factor graph of Spline BA (optimization of only the control points). For simplicity a unique node factor is considered per frame (color coded, \tikzcircle[fill=black]{2pt} - \tikzcircle[blizzardblue, fill=blizzardblue]{2pt}). Each observation can influence three node variables. Fixing state variables (the ones without edges) deals with the gauge freedoms \cite{triggs1999bundle}.}
    \label{fig:fig4}
\end{figure}

Our method adds a new control point at each image timestamp $t_i$. 
This has cost benefits, since $t=t_i$ at Eq. \ref{eq:Tinter_cumul2}, the control point $\bT_i$ has no influence during the optimization (see how in Fig. \ref{subfig:cumul_basis}, $\tilde{B}_i(t_i)=0$). However, it slightly reduces the interpolation capability. We argue that this loss is not significant since this control point has the smallest weight $\tilde{B}_i(t)$ during the interpolation, as inferred with Eq. \ref{eq:cumul_basis}.

Since the \red{time spacing between control knots (timestamps of images) is quasi-constant}, at $t_i$, the control point with greater influence is $\bT_{i-2}$. This can intuitively be seen in Fig. \ref{subfig:cumul_basis}. From this observation, at $t_i$ we initialize the origin of $\bT_{i-2}$ to the centroid of the observed object point cloud. Its orientation is initialized with the one of the previous control point. Because of this protocol, to compute the interpolation $\bT_{wo}(t_i)$, we need to have estimations of $\{\bT_j\}_{j=i-3}^{i}$, so we wait until we have 4 observations to optimize its trajectory. 

A simple factor graph for Spline BA with this protocol is shown in Fig. \ref{fig:fig4}. For each set of observations in one image, we only add one control point to the optimization. This is done to deal with the gauge freedoms \cite{triggs1999bundle} that arise from optimizing a unique pose, $\bT_{wo}$, with another 4 poses (the control points). \red{Specifically, we fix the first (and second, in Local BA) and last two control points in the sliding window, thereby fixing the most optimized variables and the ones supported by the fewest number of observations respectively.}

\subsection{Front-end}\label{subsec:frontend}

To bootstrap the trajectory of an object, we first extract $N$ (100 in our experiments) Shi-Tomasi features \cite{shi1994good} in the image region covered by one free mask, creating a point cloud used to initialize the first control point with its centroid and principal axes and subsequently initialize the set of 3D object points $\bp_o$. Feature tracking is done with KLT \cite{bouguet2001pyramidal}. A mask is considered \textit{free} if it has not already been associated with a tracked object. An association is done when the major part of the tracked object features lie on a mask.

\begin{figure}[t]
    \centering
    \begin{subfigure}[t]{0.49\linewidth}
      \centering
      \includegraphics[width=1.0\textwidth]{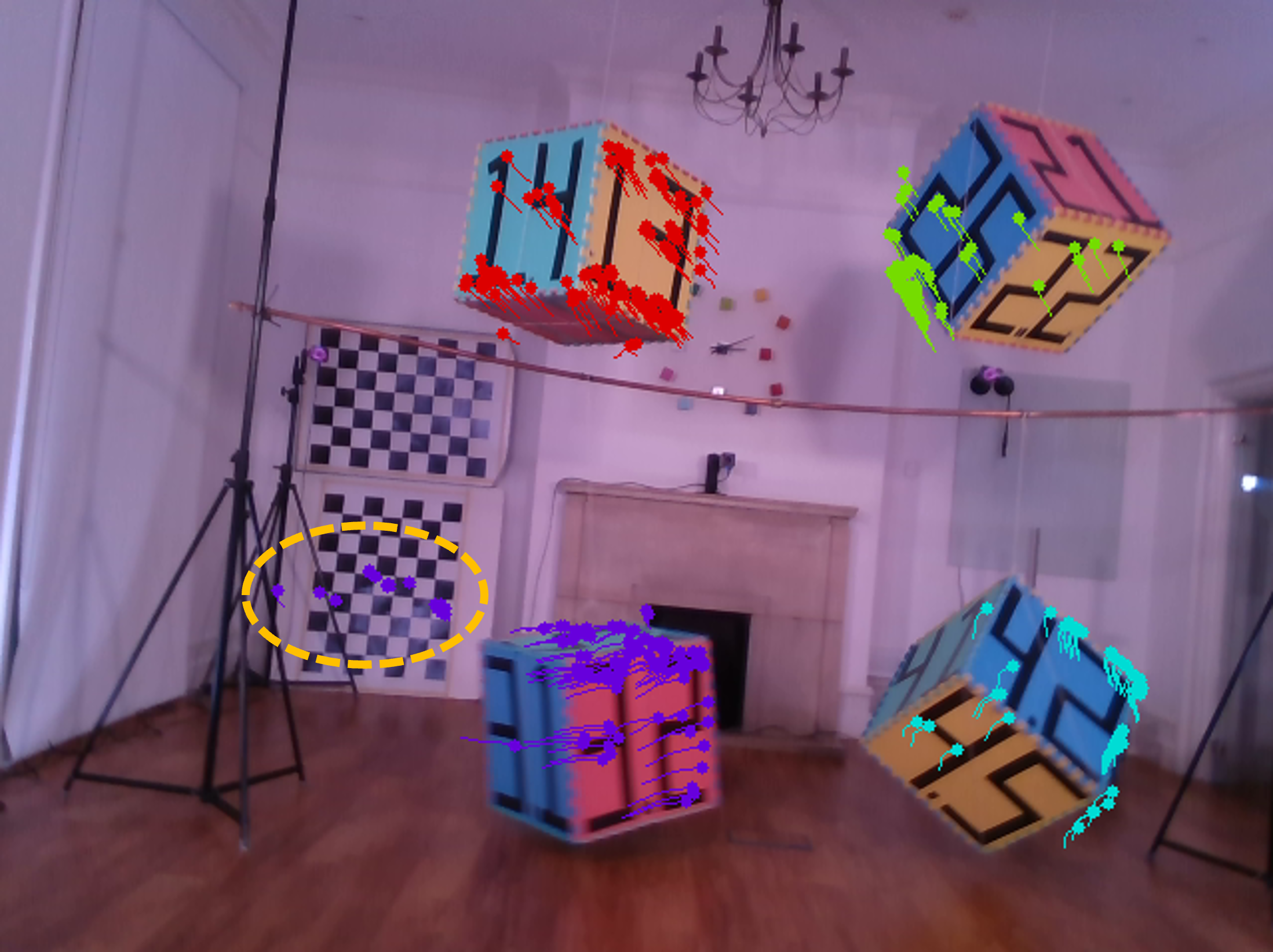}
        \caption{}
      \label{subfig:fig5a}
    \end{subfigure}
    \hfill
    \begin{subfigure}[t]{0.49\linewidth}
      \centering
      \includegraphics[width=1.0\textwidth]{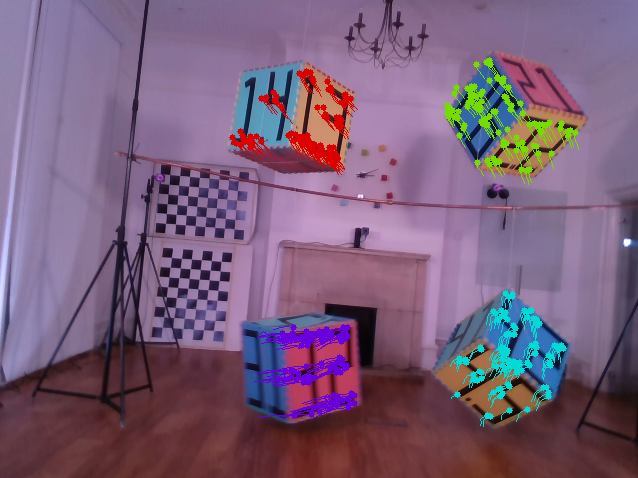}
        \caption{}
      \label{subfig:fig5b}
    \end{subfigure}
    \caption{Feature tracks, shown as circles with lines indicating previous locations, obtained during a sequence from \cite{judd2019oxford}. (a) With naive tracking, features tend to accumulate at the object borders and spurious tracks (see dashed yellow circle) appear. (b) Improvements in feature tracking after imposing backward consistency check and mask refinement.}
    \label{fig:fig5}
    \vspace{-0.5cm}
\end{figure}

When tracking of features stop being successful, new ones are extracted, aiming at keeping its number at $N$. Since at the current timestamp $t_i$ we have not initialized yet the control points $\bT_{i-1}$ and $\bT_{i}$, we cannot compute the interpolation $\bT_{wo}(t_i)$. To tackle this problem, the extracted features at time $t_i$ are tracked backwards to the image at $t_{i-2}$, at which an estimation of $\bT_{wo}(t_{i-2})$ is available and hence we can estimate its 3D coordinates $\bp_o$.

Applying naively the previous tracking method would lead to bad performance, as shown in Fig. \ref{subfig:fig5a}. On the one hand, once features are tracked to the border of an object, they tend to accumulate there since a region that contains the background has more photometric similarity. On the other hand, since masks are not perfect, features might be extracted at regions which do not belong to the object. For these reasons we impose backward consistency in the optical flow with the previous frame (up to $\sim0.1$ pix.) and a mask refinement that discards pixels whose depth value strongly deviates from the robust median absolute value (MAD) \cite{huber2004robust} of the mask depths. A qualitive comparison between the naive and this improved tracking is shown in Fig. \ref{fig:fig5}.

\section{EXPERIMENTAL RESULTS}

\subsection{Jacobian computations}\label{sec:exp_timings}

\begin{wraptable}{r}{3.5cm}
    \vspace{-0.3cm}
    \red{
    \begin{tabular}{l|r}
        \hline
        Method    & Time  \\ \hline
        \textbf{Ana.}  & $\mathbf{2.51}$ \\
        \textbf{Ana. (Lie)}  & $\mathbf{2.84}$ \\
        Num.   & $43.24$     \\
        Num. (Lie) & $50.42$ \\
        Auto. & $126.43$    \\ 
        Auto. (Lie) & $60.38$ \\ \hline
    \end{tabular}
    }
    \caption{\red{time [ms] for computing the same Jacobian.}}
    \label{wrap-tab:tab_timing}
\end{wraptable}
Table \ref{wrap-tab:tab_timing} shows a comparison between running times for computing the Jacobian of an interpolated pose\red{, $\bT$,} with respect to its 4 control points. They are obtained
with an Intel i5-7400 CPU (throughout all experiments) in Python. The analytical \red{(Ana.)} method corresponds to the Jacobians derived in Sec. \ref{subsec:jacder}. Numerical differentiation \red{(Num.)} is computed via central differences. Autograd \cite{maclaurin2015autograd} is used for automatic differentiation \red{(Auto.). If ``(Lie)'' is specified then $\Log(\bT)$ is differentiated ($\bT_{\text{vec}}$ otherwise). Our derivations reduce the time execution by at least one order of magnitude.}
\begin{figure}[b]
    \centering 
    \includegraphics[width=0.4\textwidth]{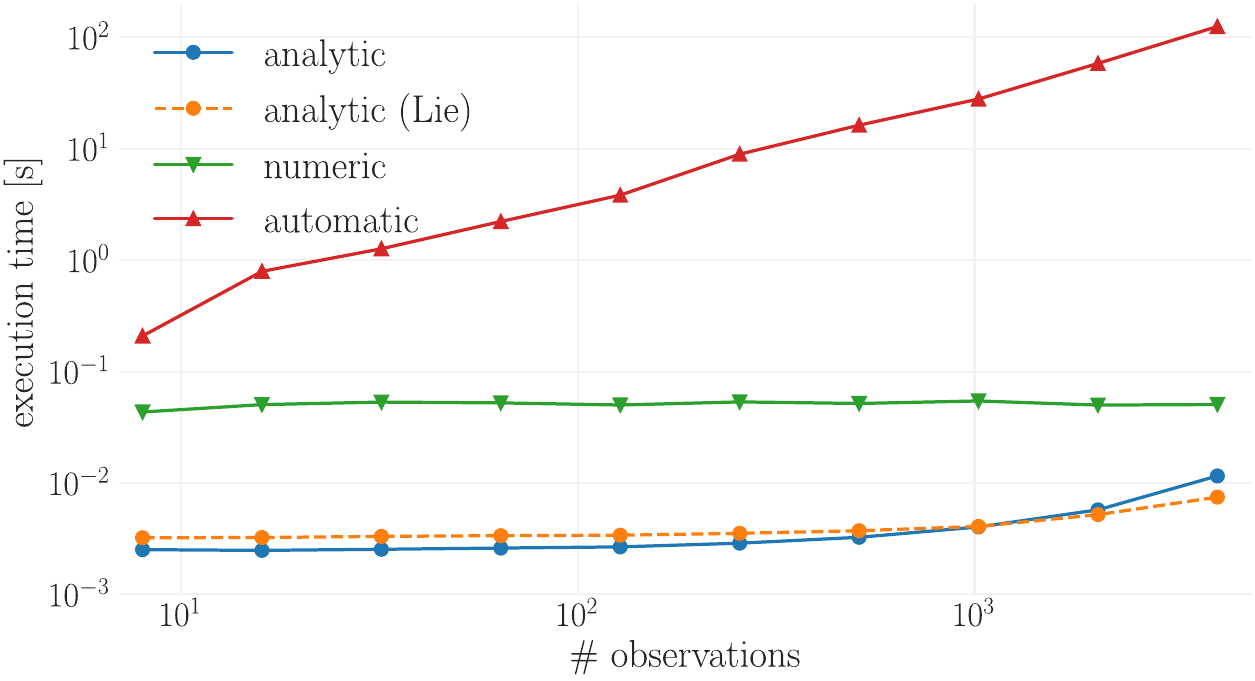}
    \caption{\red{Jacobian computation timings whith Eqs. \ref{eq:aux_jac} included.}}
    \label{fig:timing}
\end{figure}
\begin{figure*}[t]
    \centering
    \begin{subfigure}[t]{0.24\linewidth}
      \centering
     \includegraphics[height=6cm]{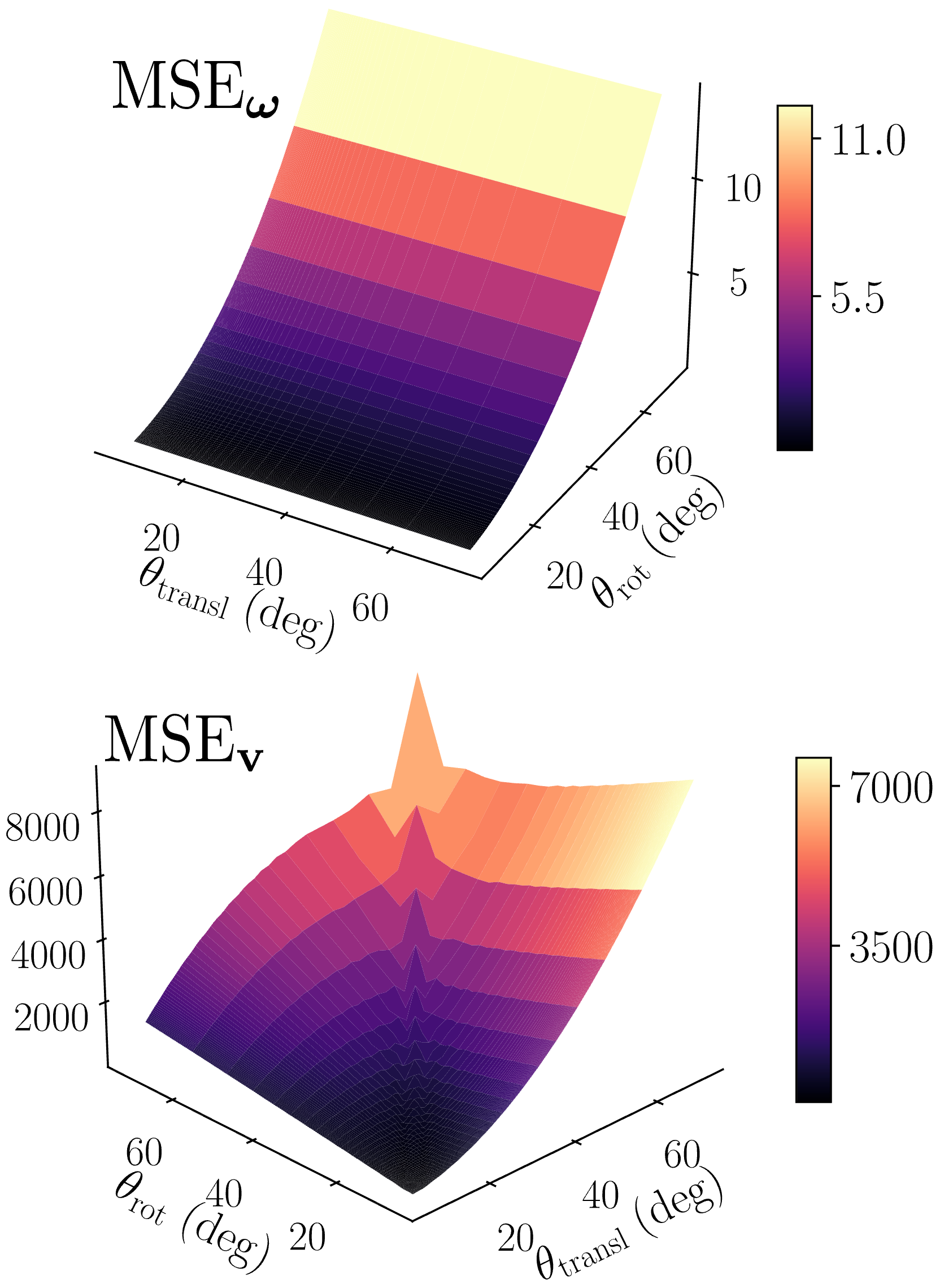}
      \caption{Discrete-time (coupled).}
    \end{subfigure}
    \hfill
    \begin{subfigure}[t]{0.24\linewidth}
      \centering
      \includegraphics[height=6cm]{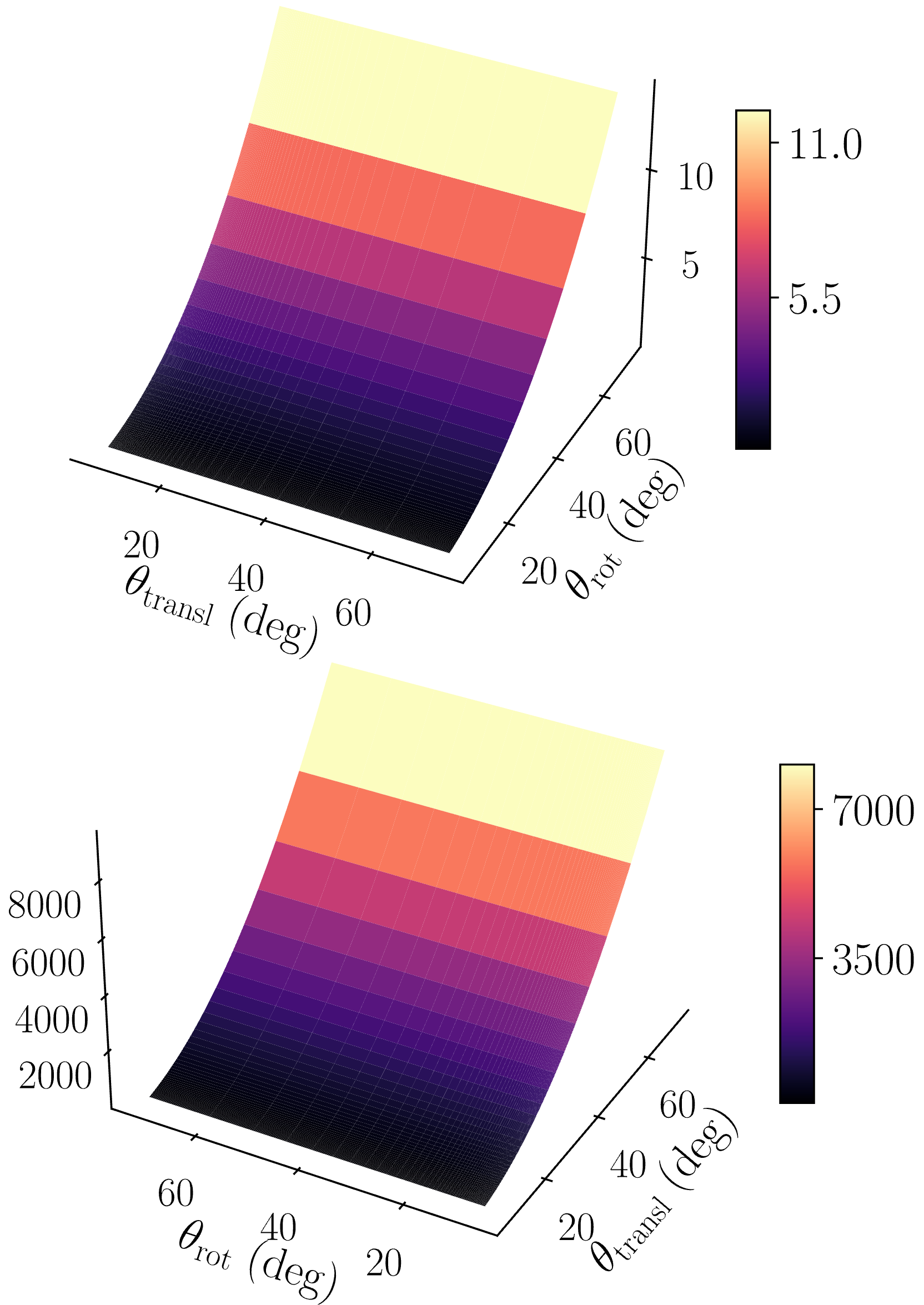}
      \caption{Discrete-time (decoupled).}
    \end{subfigure}
    \hfill
    \begin{subfigure}[t]{0.24\linewidth}
      \centering
      \includegraphics[height=6cm]{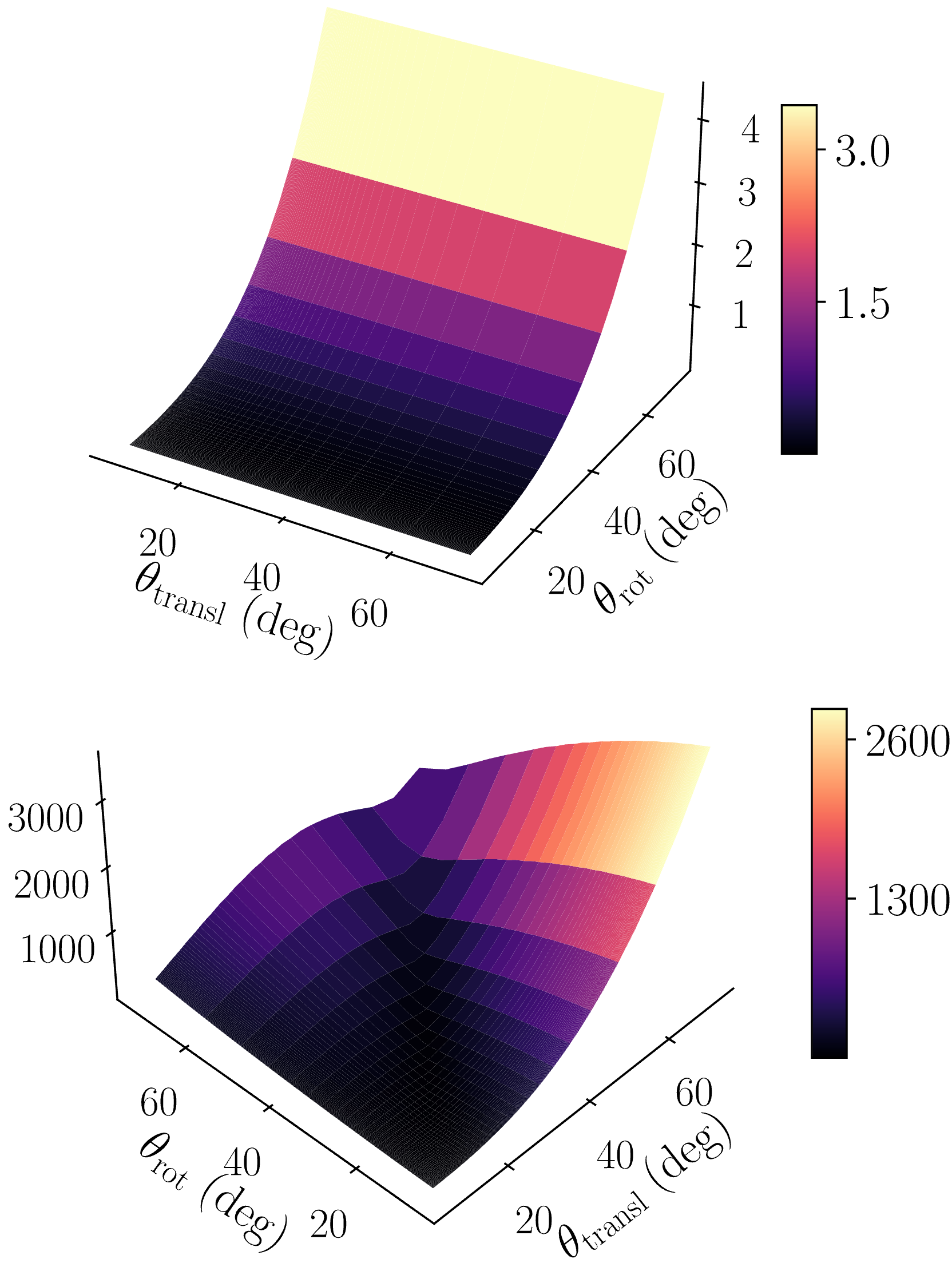}
      \caption{Continuous-time.}
    \end{subfigure}
    \hfill
    \begin{subfigure}[t]{0.24\linewidth}
      \centering
    \includegraphics[height=5cm]{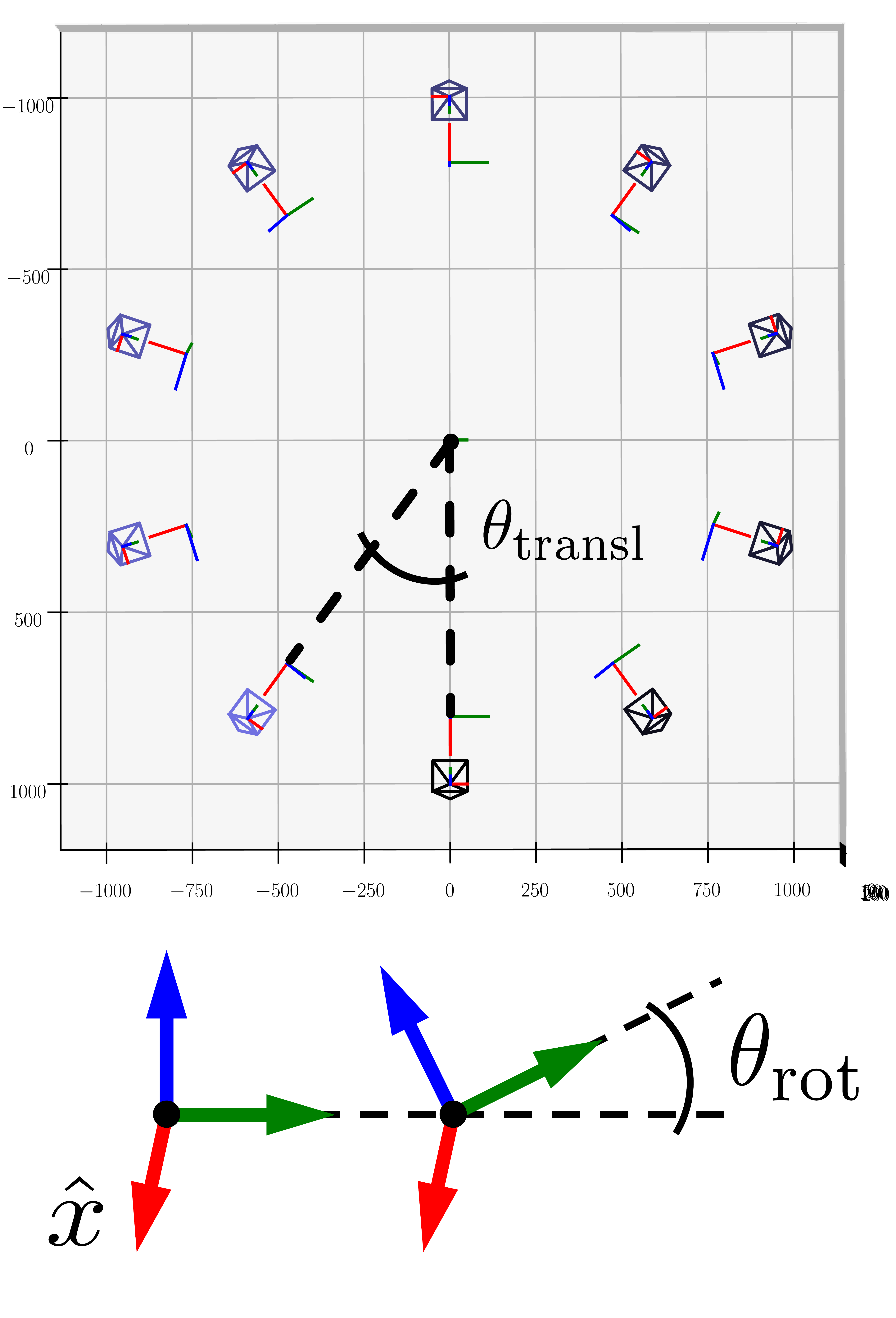}
      \caption{Schematic of the experiment.}
      \label{subfig:sch_vel}
    \end{subfigure}
    \caption{(a), (b), (c): MSE surfaces of angular [(rad/s)$^2$] (top) and linear [(mm/s)$^2$] (bottom) velocity estimations for a circular trajectory parameterized with $\theta_{\text{transl}}$ and $\theta_{\text{rot}}$ (spatial and orientation increment between timestamps) defined as in (d). (a) and (b) correspond with Eqs. \ref{eq:coupled} and \ref{eq:decoupled}. Observe how the continuous-time estimation yields significantly lower errors.}
    \label{fig:velocity}
    \vspace{-0.2cm}
\end{figure*}

Since our particular end goal is computing the Jacobians of the observation errors with respect to the control points ($\bJ_{\be_{\bp_c}} = \partial\be_{\bp_c}/\partial\bx$), correspondent timings for a varying number of observations are also shown in Figure \ref{fig:timing}, where:
\begin{align}
    \frac{\pt\be_{\bp_c}}{\pt(\bT_{wo})_{\vect}}&=-\tilde{\bp}_o^\top\otimes\bR_{cw},\label{eq:aux_jac} \\
    \red{\frac{\pt\be_{\bp_c}}{\pt\Log(\bT_{wo})}}&\red{=-\begin{bmatrix} \mathbf{I}_{3\times3} & -\bp_o^{\wedge} \end{bmatrix} \bJl(\text{Ad}_{\bT_{cw}}\Log(\bT_{wo}))\text{Ad}_{\bT_{cw}},} \nonumber
\end{align}
\red{are} the additional Jacobian\red{s} composed with the chain rule. \red{Both} analytical fashion\red{s} ha\red{ve} better performance than the alternatives. In our implementation, since $N\sim 100$, the improvement is at least of one magnitude order.

\subsection{Velocity estimations}

We evaluate linear and angular velocity estimation errors with both continuous-time (CT) and discrete-time (DT) formulations in a synthetic setup. The evaluated trajectories (see Fig. \ref{subfig:sch_vel}) consist of an object following a global z-axis turn parameterized with an angle $\theta_{\text{transl}}$ and a rotation over its own x-axis parameterized with the angle $\theta_{\text{rot}}$. Both angles measure the relative increment between consecutive timestamps $t_k$ and $t_{k+1}$. A wide range of $\{\theta_{\text{transl}}, \theta_{\text{rot}}\}$ is used to evaluate both small and big increments.

For DT, as it is common in the literature \cite{zhang2020vdo, bescos2021dynaslam}, we assume constant kinematics between two timestamps, with $t_{k+1}=t_k+\Delta t$. The linear ($\bv_o$) and angular ($\bomega_o$) velocity at $t\in[t_k,t_{k+1})$, for \textit{coupled} and \textit{decoupled} translation and rotation, are given by Eqs. \ref{eq:coupled} and \ref{eq:decoupled} respectively.
\begin{alignat}{2}
    &\btau_{o,k} = 
    \begin{bmatrix}
    \bv_{o,k}^\top & \bomega_{o,k}^\top
    \end{bmatrix}^\top =
    \frac{1}{\Delta t} \Log(\bT_{wo,k}^{-1}\bT_{wo,k+1}), \label{eq:coupled}\\
    &\bv_{o,k} = \frac{1}{\Delta t} \bR_{wo,k}^\top \bt_{w_{k,k+1}}, \,\,
    \bomega_{o,k} = \frac{1}{\Delta t} \Log(\bR_{o_{k,k+1}}), \label{eq:decoupled}
\end{alignat}
with $\bt_{w_{k,k+1}}= \bt_{wo,k+1}-\bt_{wo,k}$, $\bR_{o_{k,k+1}} = \bR_{wo,k}^\top\bR_{wo,k+1}$.

Since we are only interested in evaluating the velocity estimations, we assume a known object location (best case). Thereby, for the DT evaluation, the object reference frames match the ground-truth. For CT, since there is no closed form solution for the control points given a trajectory, we also match them to the object poses. Although this disadvantages CT estimations, it can be seen in Fig. \ref{fig:velocity} that they still yield the lowest mean squared errors MSE$_{\bv}$ and MSE$_{\bomega}$.

\subsection{Comparison against baselines}

Finally, we evaluate the whole method in the sequence swinging\_4\_unconstrained of OMD \cite{judd2019oxford}, which contains Vicon ground truth trajectories of 4 textured boxes experimenting multiple and independent $SE(3)$ motions. Since our system estimates the control points of a cumulative B-Spline curve, for this evaluation we compute the interpolation $\bT_{wo,k}(t_k)$ at each timestamp $t_k$ of the sequence with Eq. \ref{eq:Tinter_cumul2}. Following previous works, two $SE(3)$ transformations are applied to align both the global and object coordinate systems by only using the first 50 images.

In Table \ref{tab:tabMVO} we compare our system against the state of the art using the metrics reported in \cite{huang2020clustervo, judd2021multimotion}, which are the maximum component of the translation error and the norm (instead of each component) of the maximum angular errors. We compare with the pose-only results of \cite{judd2021multimotion} since it resembles the most to our optimization (we do not include kinematics in the error term). Additionally in Table \ref{tab:ate}, we compare the Absolute Trajectory Error (ATE), which measures the global consistency, against  \cite{bescos2021dynaslam}.

\begin{table}[bt]
\centering
\resizebox{0.49\textwidth}{!}{%
\begin{tabular}{l|cc|cc|cc|cc}
\hline
\multicolumn{1}{c|}{\multirow{2}{*}{System}} & \multicolumn{2}{c|}{Box 1}   & \multicolumn{2}{c|}{Box 2}   & \multicolumn{2}{c|}{Box 3}   & \multicolumn{2}{c}{Box 4}    \\
\multicolumn{1}{c|}{}                        & xyz    & $\lVert$rpy$\lVert$ & xyz    & $\lVert$rpy$\lVert$ & xyz    & $\lVert$rpy$\lVert$ & xyz         & $\lVert$rpy$\lVert$ \\ \hline
\cite{judd2021multimotion} MVO (Pose)                     & $\mathbf{0.09}$ & $11.21$    & $0.31$ & $68.20$             & $\mathbf{0.13}$ & $\mathbf{5.40}$ & $0.55$ & $93.14$                    \\
\cite{huang2020clustervo}  ClusterVO                          & $0.24$ & $\mathbf{6.09}$     & $0.45$ & $\mathbf{66.70}$    & $0.24$ & $15.03$                  & $4.69$ & $193.54$                   \\
\cite{zhang2020vdo}        VDO                           & $1.06$ & $56.77$             & $0.40$ & $169.58$            & $1.30$ & $19.12$                  & $0.76$ & $155.55$                   \\
\textbf{Ours} (w/ Local BA)                           & $0.39$ & $42.27$             & $\mathbf{0.30}$ & $126.11$   & $0.77$ & $33.33$                  & $0.47$ & $169.92$                   \\
\textbf{Ours} (w/o Local BA)                          & $0.29$ & $36.55$             & $0.38$ & $107.28$            & $0.27$ & $26.44$                  & $\mathbf{0.32}$ & $\mathbf{61.28}$  \\ \hline
\end{tabular}
}
\caption{Maximum component of translation error [m] (xyz), and norm of the maximum angular error [deg] ($\lVert$rpy$\lVert$), in swinging\_4\_unconstrained sequence \cite{judd2019oxford}.}
\label{tab:tabMVO}
\end{table}

\begin{table}[t]
\centering
\begin{tabular}{l|cccc}
\hline
System              & Box 1  & Box 2  & Box 3  & Box 4  \\ \hline
\cite{bescos2021dynaslam} DynaSLAM II         & $0.41$          & $0.37$          & $1.09$          & $0.28$ \\
\textbf{Ours} (w/ Local BA)  & $0.16$          & $\mathbf{0.18}$ & $0.37$          & $0.38$ \\
\textbf{Ours} (w/o Local BA) & $\mathbf{0.12}$ & $0.19$          & $\mathbf{0.12}$ & $\mathbf{0.21}$ \\ \hline
\end{tabular}
\caption{ATE [m] in swinging\_4\_unconstrained \cite{judd2019oxford}.}
\label{tab:ate}
\vspace{-0.5cm}
\end{table}

In terms of the trajectories global consistency, our system consistently gives lower errors than \cite{bescos2021dynaslam}. We believe this is due to its constant velocity assumption between frames. Our system has also the flexibility of estimating a constant velocity but is not constrained to only that, it can exploit more complex kinematics due to its $\mathbb{C}^2$ continuity. 

In terms of the maximum translational error, our proposal gives lower values in at least half of the motions. However, it only performs better in one of the trajectories w.r.t. the angular errors. We believe this is due to reaching local minima during the optimization. This situation is specially harmful, since this  propagates to several timestamps due to the interpolation nature of our formulation. We think that this can be addressed with a more sophisticated optimization.

\section{CONCLUSIONS AND FUTURE WORK}

This work presents a continuous-time 6-DoF tracking approach for dynamic objects observed by a mobile RGB-D sensor. This is done by fitting their trajectories to cubic cumulative B-Spline curves.
Special care has been taken in reducing the computational costs by deriving the analytical Jacobians of the interpolated pose with respect to the control points, thus promoting real-time capabilities for future works using this kind of curve. 
The evaluation has shown the potential of the proposal. Our results are on par with the state of the art, showing significant improvements in certain aspects like global consistency and velocity estimation.

As future work, we find interesting to explore higher order continuity curves. This could increase the flexibility of the trajectories as the recent work \cite{9682589} suggests. Additionally, integration with a real-time SLAM system can increase its applicability, something that could be achieved thanks to our sequential formulation and analytical Jacobians. Finally, to discover new objects in the scene, we find motion clustering \cite{judd2018multimotion} very promising instead of relying on 2D masks.









{
\balance
\bibliographystyle{IEEEtran}
\bibliography{biblio}
}


\begin{appendices}

\section{Extending Jacobians w.r.t. control points to non-cubic cumulative B-Spline curves}

The Jacobians derived in Section \ref{subsec:jacder} are designed to be used with a \textit{cubic} cumulative B-Spline curve (degree $k=4$). This type of curve is commonly used in the literature \cite{lovegrove2013spline, patron2015spline, kerl1dense, mueggler2018continuous} and it is also used in our tracking method. However, we can easily extend these Jacobians to other degrees.

To do so, we first generalize Eq. \ref{eq:app1} to a degree $k>1$:
\begin{alignat}{2}
    &\bT(t) &&= \Exp(\ba_0)\bT_0\prod_{j=1}^{k-1}\bA_j(t),
\end{alignat}
Eqs. \ref{eq:app2}-\ref{eq:app3} remain the same as they are defined for individual terms. Now, the needed applications of the multi-variable chain rule can be generalized by only changing the boundaries of $j$:
\begin{alignat}{3}
    &\bxi_{j\in\{0\,..\,k-2\}} &&: 
    \frac{\pt\bT_{\vect}}{\pt\bxi_j}(\bzero) = &&
    \frac{\pt\bT_{\vect}}{\pt \ba_j}\frac{\pt\ba_j}{\pt\bxi_j}(\bzero) + \label{eq:derapp1}\\
    &&&&&\frac{\pt\bT_{\vect}}{\pt \ba_{j+1}}\frac{\pt\ba_{j+1}}{\pt\bxi_j}(\bzero),\nonumber\\
    &\bxi_{j\in\{0\,..\,k-2\}} &&: 
    \frac{\pt\Log(\bT)}{\pt\bxi_j}(\bzero) = &&
    \frac{\pt\Log(\bT)}{\pt \ba_j}\frac{\pt\ba_j}{\pt\bxi_j}(\bzero) + \\
    &&&&&\frac{\pt\Log(\bT)}{\pt \ba_{j+1}}\frac{\pt\ba_{j+1}}{\pt\bxi_j}(\bzero),\nonumber\\
    &\bxi_{k-1} &&:
    \frac{\pt\bT_{\vect}}{\pt\bxi_{k-1}}(\bzero) = &&
    \frac{\pt\bT_{\vect}}{\pt \ba_{k-1}}\frac{\pt\ba_{k-1}}{\pt\bxi_{k-1}}(\bzero),\\
    &\bxi_{k-1} &&:
    \frac{\pt\Log(\bT)}{\pt\bxi_{k-1}}(\bzero) = &&
    \frac{\pt\Log(\bT)}{\pt \ba_{k-1}}\frac{\pt\ba_{k-1}}{\pt\bxi_{k-1}}(\bzero). \label{eq:derapp2}
\end{alignat}
For both the minimal representation of the $se(3)$ form, $\Log(\bT)$, and the 12-d vectorized version of the $SE(3)$ object, $\bT_{\vect}$.

The last re-definitions needed are in regards of the terms $\bP_j$ and $\bN_j'$ of Eqs. \ref{eq:extra1}-\ref{eq:extra2}. These are readily generalized by:
\begin{alignat}{3}
    &j=0 &&\quad\to\quad \bP_0=\bI_{4\times4},\\
    &j=1 &&\quad\to\quad \bP_1=\bT_0,\\
    &j\in\{2\,..\,k-1\} &&\quad\to\quad \bP_j=\bT_0\prod_{i=1}^{j-1}\bA_i,\\
    &j=0 &&\quad\to\quad \bN_0' = \bT,\\
    &j\in\{1\,..\,k-1\} &&\quad\to\quad \bN_j' = \prod_{i=j}^{k-1} \bA_i.
\end{alignat}
With this generalized notation, all terms from Eq. \ref{eq:app4} to Eq. \ref{eq:app5} remain the same since they are derived for individual terms. Therefore, Eqs. \ref{eq:derapp1}-\ref{eq:derapp2} are again completely defined by:
\begin{alignat}{3}
    &\frac{\pt\ba_k}{\pt \bxi_{k}}(\bzero) && =
    \tilde{B}_k(t)\,\,\bJ_l^{-1}(\Log(\bT_{k-1}^{-1}\bT_k))\text{Ad}_{\bT_{k-1}^{-1}},\label{eq:app_end}\\
    &\frac{\pt\bT_{\vect}}{\pt \ba_k} &&= 
    ((\bN_k')^\top \otimes \bR_{\bP_k})\,\bG\,\bJ_l(\ba_k),\\
    & \frac{\pt\Log(\bT)}{\pt\ba_k}&&=\bJ_l^{-1}(\Log(\bT))\text{Ad}_{\bP_k}\bJ_l(\ba_k).
\end{alignat}

\section{Jacobians of velocity in $SE(3)$}

In this appendix we show how to extend to $SE(3)$ the Jacobian of the velocity vector, $\btau_o^{(k)}$, with respect to the control points.
It is derived for cumulative B-Splines of degree $k$ on $SO(3)$ in \cite{sommer2020efficient}. This vector is expressed in the local coordinate system $\{o\}$ and, in our case, contains not only the angular, but also the linear velocity of the interpolated motion with a cumulative B-Spline of degree $k$ ($k=4$ when cubic).

Concretely, in \cite[Th.~5.2]{sommer2020efficient} is shown that this vector can be constructed in the following recursive manner, for $j\in\{1\,..\,k\}$ (and following our notation):
\begin{alignat}{3}
    &\btau^{(j)} && = \text{Ad}_{\bA_{j-1}^{-1}}\btau^{(j-1)} + \dot{\tilde{B}}_{j-1}\bOmega_{j-1},\label{eq:final2}\\
    &\btau^{(1)} &&  = \bzero.
\end{alignat}
Therefore, the Jacobian of $\btau^{(j)}$ w.r.t. $\bOmega_j$ is zero. 

Following similar steps as in \cite{sommer2020efficient}, we can first compute the Jacobian $\pt\btau^{(j+1)}/\pt\bOmega_j$ and later generalize it to $\pt\btau^{(k)}/\pt\bOmega_j$:
\begin{equation}
    \frac{\pt\btau^{(j+1)}}{\pt\bOmega_j} = 
    \frac{\pt\text{Ad}_{\bA_{j-1}^{-1}} \btau^{(j-1)}}{\pt\bOmega_j} + \dot{\tilde{B}}_{j-1}\mathbf{I}_{6\times6} \label{eq:app_final}
\end{equation}
where:
\begin{alignat}{3}
    &\frac{\partial\text{Ad}_{\bA^{-1}_{j-1}}\btau^{(j-1)}}{\partial \bOmega_{j-1}} \overset{(\ref{eq:app2})}{=} 
    \frac{\partial\text{Ad}_{\Exp(-\ba_{j-1})}\btau^{(j-1)}}{\partial \bOmega_{j-1}}\\
    & \overset{(\ref{eq:app2b})}{=} 
    \tilde{B}_{j-1} \frac{\partial\text{Ad}_{\Exp(-\ba_{j-1})}\btau^{(j-1)}}{\partial \ba_{j-1}}\\
    &\overset{(\ref{eq:Jl1})}{=} \tilde{B}_{j-1} \frac{\partial \text{Ad}_{\Exp(-\bJ_l(-\ba_{j-1})\bxi)\Exp(-\ba_{j-1})}\btau^{(j-1)}}{\partial \bxi}(\bzero) \\
    &\overset{\text{\cite[Eq.~34]{sola2018micro}}}{=} \tilde{B}_{j-1} \frac{\partial \text{Ad}_{\Exp(-\bJ_l(-\ba_{j-1})\bxi)}\text{Ad}_{\Exp(-\ba_{j-1})}\btau^{(j-1)}}{\partial \bxi}(\bzero) \\
    & = -\tilde{B}_{j-1}\frac{\partial \text{Ad}_{\Exp(\bxi)}\text{Ad}_{\Exp(-\ba_{j-1})}\btau^{(j-1)}}{\partial \bxi}(\bzero)\,\,\bJ_l(-\ba_{j-1})
\end{alignat}
Now, by compacting the expression, denoting $\mathbf{h}_{j-1}=[\mathbf{h}^\top_v, \mathbf{h}^\top_w]^\top$ (with $\mathbf{h}_{j-1}\in\mathbb{R}^6$, and both $\mathbf{h}^\top_v, \mathbf{h}^\top_{\bomega}\in\mathbb{R}^3$) as $\text{Ad}_{\Exp(-\ba_{j-1})}\btau^{(j-1)}$, we have that:
\begin{equation}
    \frac{\partial\text{Ad}_{\bA^{-1}_{j-1}}\btau^{(j-1)}}{\partial \bOmega_{j-1}} =
    -\tilde{B}_{j-1}\frac{\partial \text{Ad}_{\Exp(\bxi)}\mathbf{h}_{j-1}}{\partial \bxi}(\bzero)\,\,\bJ_l(-\ba_{j-1}),
\end{equation}
from which, taking into account the form of the Adjoint of SE(3) \cite[Example~6]{sola2018micro}, we end up having:
\begin{equation}
    \frac{\partial\text{Ad}_{\bA^{-1}_{j-1}}\btau^{(j-1)}}{\partial \bOmega_{j-1}} = 
    \tilde{B}_{j-1}\begin{bmatrix}
    \mathbf{h}_w^\wedge & \mathbf{h}_v^\wedge\\
    \bzero_{3\times3} & \mathbf{h}_w^\wedge
    \end{bmatrix}\bJ_l(-\ba_{j-1}).
\end{equation}
Thus completing the derivation of Eq. \ref{eq:app_final}.

Using this last derivation and following \cite{sommer2020efficient}, $\pt\btau^{(k)}/\pt\bOmega_j$ can be computed in a recursive manner (from $j=k-1$ to $j=1$):
\begin{alignat}{3}
    &\mathbf{P}_{k-1} &&= \mathbf{I}_{6\times6},\\
    &\frac{\pt\btau^{(k)}}{\pt\bOmega_j} &&= \mathbf{P}_j \frac{\pt\btau^{(j+1)}}{\pt\bOmega_j},\label{eq:app_rec}\\
    &\mathbf{P}_{j-1} &&= \mathbf{P}_{j} \text{Ad}_{\bA_{j-1}^{-1}}
\end{alignat}
The proof of this lies by iterative differentiation of Eq. \ref{eq:final2}. From \ref{eq:app_final}, we already know $\pt\btau^{(k)}/\pt\bOmega_{k-1}$, thus e.g. for $j=k-2$:
\begin{equation}
    \frac{\pt\btau^{(k)}}{\pt\bOmega_{k-2}} = \text{Ad}_{\bA_{k-1}^{-1}} \frac{\pt\btau^{(k-1)}}{\pt\bOmega_{k-2}}
\end{equation}
since $\text{Ad}_{\bA_{k-1}^{-1}}$ does not depend on $\bOmega_{j}$ for $j\leq k-2$. Thereby, differentiating the same expression for the successive values of $j\leq k-2$, leads to the recursive scheme of Eq. \ref{eq:app_rec}.

Lastly, since we know $\pt\btau^{(k)}/\pt\bOmega_{j}$, we just need to compose this derivative, via the chain rule, with the derivative $\pt\bOmega_{j}/\pt\bxi_j$. We already now this from Eq. \ref{eq:app_end}, since:
\begin{equation}
    \frac{\pt\ba_j}{\pt\bOmega_j} = \tilde{B}_j,
\end{equation}
thereby:
\begin{equation}
    \frac{\pt\bOmega_j}{\pt \bxi_{j}}(\bzero) =
    \bJ_l^{-1}(\Log(\bT_{k-1}^{-1}\bT_k))\text{Ad}_{\bT_{k-1}^{-1}}.
\end{equation}
Thus completing the Jacobian derivation of $\btau^{(k)}\in\mathbb{R}^6$ (minimal representation of $se(3)$) w.r.t. the control points.

\end{appendices}

\end{document}